\documentclass{article}

\PassOptionsToPackage{numbers, compress}{natbib}
\usepackage[final]{neurips_2025}

\usepackage[utf8]{inputenc} 
\usepackage[T1]{fontenc}    
\usepackage{hyperref}       
\usepackage{url}            
\usepackage{booktabs}       
\usepackage{amsfonts}       
\usepackage{nicefrac}       
\usepackage{microtype}      
\usepackage{xcolor}         

\usepackage{graphicx, amsmath, multirow, wrapfig, float}
\def\eg{\emph{e.g.}}
\def\ie{\emph{i.e.}}

\title{Normal-Abnormal Guided Generalist \\ Anomaly Detection}

\author{%
  Yuexin Wang$^{1,2}$\thanks{Equal Contribution.}\;\,, Xiaolei Wang$^{1,2*}$, Yizheng Gong$^{1,2}$, Jimin Xiao$^{1}$\thanks{Corresponding Author.}\\
  $^1$Xi'an Jiaotong-Liverpool University \\
  $^2$University of Liverpool
}

\begin{document}

\maketitle

\begin{abstract}
\label{sec:abstract}
Generalist Anomaly Detection (GAD) aims to train a unified model on an original domain that can detect anomalies in new target domains. Previous GAD methods primarily use only normal samples as references, overlooking the valuable information contained in anomalous samples that are often available in real-world scenarios. To address this limitation, we propose a more practical approach: normal-abnormal-guided generalist anomaly detection, which leverages both normal and anomalous samples as references to guide anomaly detection across diverse domains. We introduce the Normal-Abnormal Generalist Learning (NAGL) framework, consisting of two key components: Residual Mining (RM) and Anomaly Feature Learning (AFL). RM extracts abnormal patterns from normal-abnormal reference residuals to establish transferable anomaly representations, while AFL adaptively learns anomaly features in query images through residual mapping to identify instance-aware anomalies. Our approach effectively utilizes both normal and anomalous references for more accurate and efficient cross-domain anomaly detection. Extensive experiments across multiple benchmarks demonstrate that our method significantly outperforms existing GAD approaches. This work represents the first to adopt a mixture of normal and abnormal samples as references in generalist anomaly detection. The code and datasets are available at~\url{https://github.com/JasonKyng/NAGL}.
\end{abstract}

\section{Introduction}
\label{sec:intro}
\begin{figure}[tp]
\centering
\includegraphics[width=\linewidth]{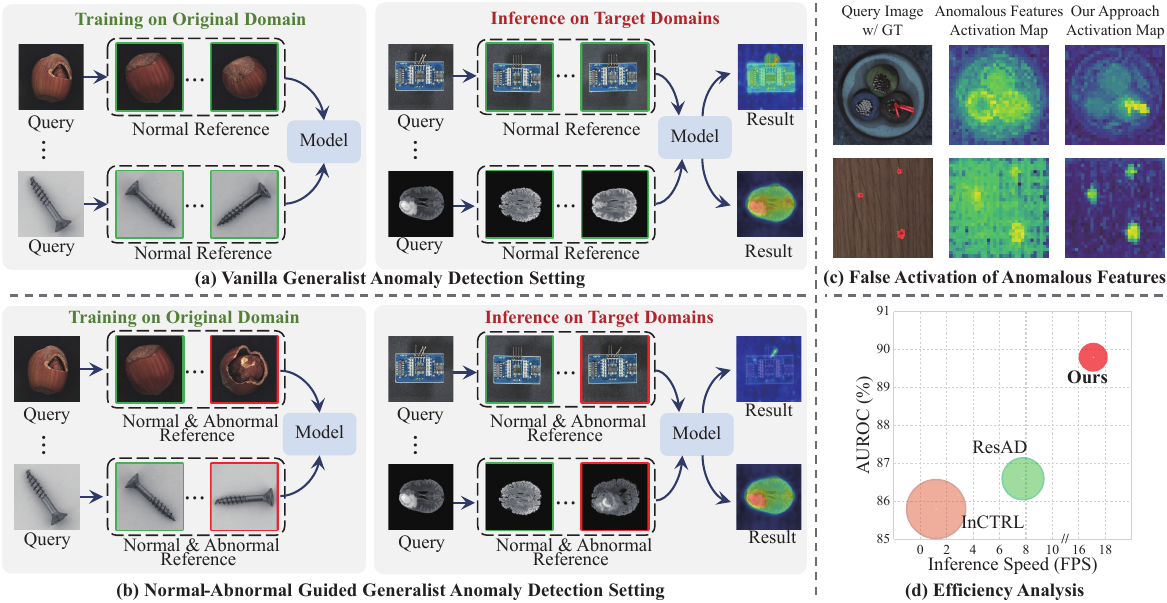}
\caption{Overview of existing and proposed GAD paradigms. (a) Vanilla GAD only adopts a few normal samples as references. (b) Our approach combines normal and abnormal references to enhance detection. (c) Direct application of the KNN-based method to our normal-abnormal guided detection task causes to false activations (middle), while our approach eliminates them (right). (d) Comparisons of different methods in terms of AUROC sample (y-axis), inference speed  (x-axis), and model size (circle radius). Our approach achieves the \textbf{highest} AUROC sample metric for anomaly detection while being $\textbf{2}\times$ \textbf{faster} than ResAD and $\textbf{14}\times$ \textbf{faster} than InCTRL.}
\label{fig:fig1}
\end{figure}

Visual Anomaly Detection (AD) \cite{PaDiM, Li_2021_CVPR, zhang2023prototypical, cao2023anomaly, ding2022catching, zhu2024anomaly, wang2025icc, yang2024promptable, guo2025dinomaly, guo2023recontrast, luo2025INP-Former, 10884560} plays a crucial role in industrial quality inspection \cite{MVTec, Roth_2022_CVPR, VisA, jiang2025mmad} and medical diagnosis \cite{ding2022catching, he2023transformers}. Its primary objectives are to classify images as normal or anomalous and to localize anomalies within those images. Traditional AD methods \cite{li2024promptad, chen2022deep, fang2023fastrecon, wang2024cnc, chen2024filter} focus on training and testing a model on a single domain, without considering how detection capabilities might transfer to a different target domain. However, many real-world AD scenarios prohibit training on the target domain due to data scarcity and privacy issues, making it difficult to achieve the desired outcome in that target domain. To address the challenge, InCTRL \cite{zhu2024toward} and ResAD \cite{yao2024resad} propose Generalist Anomaly Detection (GAD) that aims to train a unified model on the original domain while enabling AD on the target domain. As shown in Fig.~\ref{fig:fig1}a, the GAD framework adopts a meta-learning strategy. This strategy trains the model in the original domain to localize the anomalous regions of a query image by referring to a limited number of normal references. Subsequently, the learned ability is transferred to the new target domain.

Although previous GAD methods \cite{zhu2024toward, yao2024resad} have made substantial progress, they are not yet practical for real-world applications. Models trained exclusively on normal samples often lack the discriminative power to reliably distinguish anomalies \cite{ding2022catching, yao2023explicit}. However, real-world scenarios often provide a small number of anomalous samples (\eg, defective parts or diagnosed disease cases). These anomalous samples contain valuable information on anomaly characteristics that could be leveraged to improve detection. Given this context, we propose a more practical and effective approach: normal-abnormal-guided generalist anomaly detection (illustrated in Fig.~\ref{fig:fig1}b). This approach leverages both normal and anomalous samples as references, guiding the model to detect anomalies across diverse domains. The core of this approach is learning the relationships between a query and these references in the original domain, and then applying this learned understanding to the target domain.

Previous methods \cite{zhu2024toward, yao2024resad} leverage residuals between queries and normal references to ensure transferability, but these approaches are specifically tailored for scenarios where few-shot normal samples are provided, which cannot be adapted to the normal-abnormal guided paradigm. When provided with a reference set with both normal and abnormal samples, KNN-based approaches \cite{Roth_2022_CVPR, damm2024anomalydino} can serve as a solution, where sample regions far away from normal references and close to abnormal references are considered as abnormal. However, these KNN-based approaches are typically training-free, lacking the adaptability offered by data-driven learning. Additionally, we experimentally find that the KNN-based method suffers from false activation problems (see Fig.~\ref {fig:fig1}c). To address these limitations, we propose a Normal-Abnormal Generalist Learning (NAGL) framework, which utilizes the residual features to model the differences between normal and anomalous references. The proposed method achieves superior detection performance at faster speeds (see Fig.~\ref{fig:fig1}d), demonstrating its significant potential for practical applications.

The core of our NAGL framework consists of two parts: Residual Mining (RM) from normal-abnormal references and Anomaly Feature Learning (AFL) for query images through residual mapping. To fully explore abnormal reference patterns and maintain transferability, RM leverages normal-abnormal residuals to learn abnormal reference patterns through a designed attention operation, obtaining residual proxies. AFL adaptively learns abnormal features of a query image by comparing residual proxies with residuals between the query feature and normal references, obtaining anomaly proxies. Finally, anomaly localization results are acquired by similarity computation between the query feature and anomaly proxies. Our NAGL only relies on residual training to adaptively capture the similarities and differences between query and normal-abnormal reference samples, so it can be transferred between different domains.

Leveraging RM and AFL, our proposed NAGL framework effectively achieves normal-abnormal guided generalist anomaly detection. The main contributions of this work are summarized as follows: 
\begin{itemize}
\item We propose a different generalist anomaly detection task and a corresponding dataset split. This task is the first to adopt a mixture of normal and abnormal samples as references.

\item We propose a novel Normal-Abnormal Generalist Learning framework to effectively leverage abnormal reference in GAD that adapts normal-abnormal reference residuals to mine potential anomalies in the query. 

\item Extensive experiments across multiple anomaly detection benchmarks demonstrate that our method significantly outperforms existing GAD approaches.
\end{itemize}

\section{Related Work}
\label{sec:related}
\subsection{Anomaly Detection}
Artificial intelligence techniques based on deep learning have been widely applied \cite{krizhevsky2012imagenet,long2015fully,wang2024towards}, with anomaly detection (AD) being a significant application. AD can be divided into various tasks according to real-world requirements, \eg, unsupervised AD \cite{wang2025dec,wang2025cnc,VisA,liu2023diversity,zhang2024realnet,strater2024generalad}, few-shot AD (FSAD) \cite{huang2022registration, fang2023fastrecon, lee2024text, lv2025oneforall, tao2024kernel, li2024onetonormal}, zero-shot AD (ZSAD) \cite{jeong2023winclip,zhou2024anomalyclip,cao2024adaclip}, noisy AD \cite{chen2022deep, jiang2022softpatch}, 3D AD \cite{gu2024rethinking, costanzino2024multimodal, liu2024real3d, zhou2025pointad}, and open-set AD \cite{ding2022catching, zhu2024anomaly}. Among these tasks, unsupervised AD generally adopts a one-class classification paradigm to train the detection model, \ie, the model is only trained on normal samples and can detect unseen abnormal patterns during the inference phase. Existing unsupervised methods can be divided into three main categories: reconstruction-based \cite{tien2023revisiting,lu2023hierarchical,he2024diffusion}, feature-embedding-based \cite{mcintosh2023inter,Roth_2022_CVPR,lei2023pyramidflow, xie2023pushing}, and augmentation-based \cite{zavrtanik2021draem,zhang2024realnet,lin2024comprehensive} methods. PatchCore \cite{Roth_2022_CVPR}, a simple feature-embedding-based method, firstly constructs a memory bank of normal embedding features, then introduces a nearest neighbour search to find several nearest neighbours for each test embedding feature, and computes the distance between the test embedding feature and its neighbours as an anomaly score. PatchCore can also be extended to a few-shot AD task well, where the used memory bank only requires a few normal reference samples to construct. Based on this work \cite{Roth_2022_CVPR}, FastRecon \cite{fang2023fastrecon} leverages ridge regression on normal features to quickly reconstruct test features. However, these methods only have a testing phase and detect anomalies on a single domain of data, lacking transferability. Therefore, some ZSAD \cite{zhou2024anomalyclip} and GAD \cite{yao2024resad,zhu2024toward} methods have begun to study cross-domain detection. And \cite{author2025mvrec} also proposes a general few-shot defect classification framework that addresses real-world applications of defect type identification. Building upon the GAD task, our proposed normal-abnormal guided generalist AD framework addresses cross-domain anomaly detection by effectively utilizing abnormal samples from the original domain to improve performance.

\subsection{Generalist Anomaly Detection}
FSAD is designed to identify anomalies using only a limited number of normal samples from target datasets. Existing FSAD methods can be divided into training-free-based \cite{Roth_2022_CVPR,fang2023fastrecon} and meta-learning-based methods \cite{huang2022registration,zhu2024toward,yao2024resad}. Meta-learning-based methods focus on generalizing the detection of the model to the new target domain. RegAD \cite{huang2022registration} trains a registration network using samples from seen categories, which can register samples from unseen categories and achieve cross-category anomaly detection. However, RegAD remains dependent on domain relevance between training and testing data. To achieve domain-agnostic anomaly detection, InCTRL \cite{zhu2024toward} firstly proposes a generalist anomaly detection (GAD) task and utilizes residual distance to discriminate anomalies. Subsequently, ResAD \cite{yao2024resad} applies residual features to eliminate domain dependencies to implement the GAD task. The reference set used by the vanilla GAD task only contains normal samples. However, in practical scenarios, a small number of abnormal samples can also be obtained. Therefore, we propose the normal-abnormal guided GAD task, where the reference set combines normal and abnormal samples. 

\begin{figure}[tp]
    \centering
    \includegraphics[width=\linewidth]{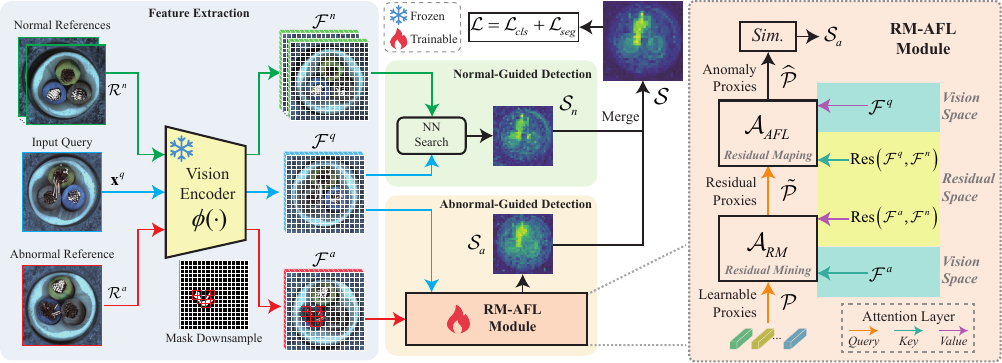}
    \caption{Overview of our proposed NAGL framework. Given a test image and its corresponding reference images (normal and abnormal), features are extracted through a pre-trained backbone network. The extracted normal features guide the generation of a normal-guided score map. Meanwhile, abnormal features are processed through the RM-AFL module to produce an abnormal-guided score map. This module implements a transformation process: from learnable proxies in vision space to residual proxies capturing normal-abnormal differences in residual space, and finally to anomaly proxies in vision space that highlight specific anomalous regions in the query image. The final anomaly score is computed by merging both normal and abnormal guided score maps.}
    \label{fig:fig3}
\end{figure}

\section{Methodology}
\label{sec:method}
\subsection{Normal-Abnormal Guided GAD Task}
In the normal-abnormal guided GAD task, we focus on training a unified model on the original domain and detecting anomalies in a new target domain using both normal and abnormal samples as references. Formally, let $\mathcal{D}_{origin}=\{\mathcal{X},\mathcal{Y}\}$ be the training dataset from original domain, where $\mathcal{X}=\{\mathbf{x}_{i}\}^{N}_{i=1}$ consists of $N$ normal and abnormal samples, $\mathcal{Y}=\{\mathbf{y}_{i}\}^{N}_{i=1}$ is corresponding ground truth masks, and each normal sample is equipped a zero mask. During training phase, we organize the data into many episodes, where each episode consists of a reference set $\mathbf{R}=\{\mathcal{R}^{n}, \mathcal{R}^{a}\}$ and a query input $\mathbf{x}^{q}$ from $\mathcal{D}_{origin}$. 
The reference set $\mathbf{R}$ contains normal samples $\mathcal{R}^{n}=\{r^{n}_{k}\}^{K_1}_{k=1}$ and abnormal samples $\mathcal{R}^{a}=\{r^{a}_k\}^{K_2}_{k=1}$, $K_{1}$ and $K_2$ denote the number of normal and abnormal reference samples, respectively. We train a unified detection model on $\mathcal{D}_{origin}$. During inference, the model is evaluated on the target domain dataset $\mathcal{D}_{target}$ containing unseen categories compared to $\mathcal{D}_{origin}$. The testing process is consistent with the training phase, where each test sample is equipped with a normal-abnormal reference set. In our proposed normal-abnormal guided GAD task, considering the scarcity of abnormal samples, we set $K_{1}\ge K_{2}$, making it highly applicable to real-world situations. 

\subsection{Overview of Proposed Method}
Our proposed Normal-Abnormal Generalist Learning (NAGL) framework is designed to detect anomalies in the target domain through original domain training. As shown in Fig.~\ref{fig:fig3}, given a query input and normal-abnormal references, we extract these features by applying a pre-trained backbone network for further processing. After feature extraction, we directly leverage the Nearest Neighbor (NN) search between the query and normal references, obtaining an initial anomaly score map. Next, to apply an abnormal reference to improve the initial anomaly map, we propose an RM-AFL module. This module consists of two attention parts: Residual Mining (RM) from normal-abnormal references and Anomaly Feature Learning (AFL) for query images with residual mapping. RM relies on normal-abnormal residuals to learn the representation for abnormal reference patterns, generating several residual proxies. AFL applies these proxies to learn abnormal areas in the query image and obtain anomaly proxies. Finally, we compute the cosine similarity between the query feature and anomaly proxies for anomaly localization results.

\subsection{Feature Extraction}
In this section, we describe the feature extraction process for an episode. For each episode, we have a normal-abnormal reference set $\mathbf{R}=\{\mathcal{R}^{n}, \mathcal{R}^{a}\}$ and a input query image $\mathbf{x}^{q} \in \mathbb{R}^{H\times W \times 3}$. For input query and references, we follow the common practice of using a pre-trained backbone network $\phi(\cdot)$ to extract their features. Subsequently, we obtain query patch features $\mathcal{F}^q=\{f^{q}_{i}\}^{L}_{i=1}$, normal reference patch features $\mathcal{F}^{n}=\{f^{n}_{i}\}^{K_{1}L}_{i=1}$, and abnormal reference patch features $\mathcal{F}^a=\{f^{a}_{i}\}^{K_{2}L}_{i=1}$, where $f^{q}_{i},f^{n}_{i},f^{a}_{i}\in \mathbb{R}^{C}$, $L=h*w$ is the number of patches in one feature map, $h$ and $w$ are length and width of feature map, and $C$ is the channel dimension of each patch feature. For abnormal reference features, we downsample the anomaly mask to match the length of $\mathcal{F}^a$, yielding $\mathcal{M}^a \in \mathbb{R}^{K_2L}$. 

\subsection{Normal-Guided Anomaly Score}
In this section, we apply query and normal reference features to obtain an initial anomaly localization result. Following the previous work \cite{Roth_2022_CVPR}, we directly apply the nearest neighbor search between query features $\mathcal{F}^q$ and normal reference features $\mathcal{F}^n$ to generate an anomaly score map $\mathcal{S}_{n}\in\mathbb{R}^L$. Specifically, we apply the NN method to find the closest reference patch feature $f^{n}_{*}$ for each query patch feature $f^{q}_{i}$ in normal reference patches $\mathcal{F}^{n}$, 
\begin{equation}\label{closet normal}
f^{n}_{*}:=\mathop{\mathbf{argmin}}_{f^{n} \in \mathcal{F}^{n}}\left(\mathbf{d}(f_i^{q},f^{n})\right),
\end{equation}
where $\mathbf{d}(\cdot,\cdot)$ is the cosine distance function and $i\in\{1,2,\cdots,L\}$. Therefore, each patch anomaly score is defined as
\begin{equation}\label{normal score map}
\mathcal{S}^{i}_{n}=\mathbf{d}(f^{q}_{i},f^{n}_{*}),
\end{equation}
where $\mathcal{S}^{i}_{n}=0$ indicates the most normal region and $\mathcal{S}^{i}_{n}=1$ indicates the most abnormal region.

\subsection{Abnormal-Guided Anomaly Score}
In this section, we focus on applying abnormal reference to improve the initial anomaly map. This section mainly consists of Residual Mining (RM) from normal-abnormal reference and Anomaly Feature Learning (AFL) for query images by residual mapping. The whole RM-AFL mainly relies on residual training to adaptively capture the similarities and differences between query and reference samples, so it can be transferred between different domains.

\subsubsection{Residual Mining from References} 
The RM module introduces a learnable query proxy to adaptively capture abnormal patterns in anomalous references and utilizes normal-abnormal residuals to learn anomalous variations through a cross-attention mechanism. Specifically, for given abnormal reference features $\mathcal{F}^a$, we employ the NN method to identify the closest normal reference patch feature for each $f_{i}^{a}$ in $\mathcal{F}^a$ and compute the residual between the two. For simplicity, we denote this process as
\begin{equation}
\mathrm{Res}(\mathcal{F}^a, \mathcal{F}^n) = \mathcal{F}^a - \mathcal{F}_{*}^{n},
\end{equation}
where $\mathcal{F}_{*}^{n}\in\mathbb{R}^{K_{2}L\times C}$ consists of $f^{n}_{*}$ corresponding to each $f^{a}_{i}$. We then design an attention layer ($\mathcal{A}_{RM}$) that utilizes normal-abnormal residuals to learn representations of abnormal variations, thereby generating residual proxies. We state \textit{Query}, \textit{Key}, and \textit{Value} in the attention computation:
\begin{equation}
\mathbf{Q}_{1}=\mathbf{W}_{1}^{\mathcal{Q}}\mathcal{P}, \mathbf{K}_1 = \mathbf{W}_1^{\mathcal{K}}\mathcal{F}^a,
\mathbf{V}_1=\mathbf{W}^{\mathcal{V}}_1\mathrm{Res}(\mathcal{F}^a, \mathcal{F}^n),
\end{equation}
where $\mathcal{P}\in \mathbb{R}^{M\times C}$ is learnable proxies and randomly initialized, parameters $\mathbf{W}_{1}^{\mathcal{Q}}$, $\mathbf{W}_1^{\mathcal{K}}$, and $\mathbf{W}^{\mathcal{V}}_1$ transform input features into query, key, and value vectors, $M$ is a hyperparameter representing the number of $\mathcal{P}$. Therefore, the attention-based RM mechanism can be described as
\begin{equation}\label{RM}
 \widetilde{\mathcal{P}} = \mathbf{SA}_1\left(\mathrm{Softmax}\left(\frac{\mathbf{Q}_1\mathbf{K}_1^\mathrm{T}}{\sqrt{d}} + \mathcal{M}^{\prime}\right)\mathbf{V}_1\right),
\end{equation}
where $\widetilde{\mathcal{P}}\in \mathbb{R}^{M\times C}$ is called residual proxies, $\mathbf{SA}_{1}(\cdot)$ denotes self-attention operation, and $d$ is a scaling factor \cite{vaswani2017attention}. The attention mask $\mathcal{M}^{'}=\alpha(1-\mathcal{M}^{a})$ contains either zero or negative infinity values, where $\alpha$ is a large negative value (\eg, $-10^9$). This mask ensures cross-attention focuses exclusively on abnormal regions. Residual proxies $\widetilde{\mathcal{P}}$ learn abnormal reference patterns and apply residuals as \textit{Value} to represent variations for anomalies.

\subsubsection{Anomaly Feature Learning for Query Images}
In this section, AFL focuses on applying obtained residual proxies $\widetilde{\mathcal{P}}$ to learn potential abnormal patterns in $\mathcal{F}^{q}$, obtaining anomaly proxies. The anomaly localization results are
acquired by similarity computation between query patch features and anomaly proxies. This process can also be described by another attention layer ($\mathcal{A}_{AFL}$). The \textit{Query}, \textit{Key}, \textit{Value} of the attention module is designed as:
\begin{equation}
\mathbf{Q}_{2}=\mathbf{W}^{\mathcal{Q}}_{2}\widetilde{\mathcal{P}}, \mathbf{K}_2 = \mathbf{W}^{\mathcal{K}}_2\mathrm{Res}(\mathcal{F}^{q},\mathcal{F}^{n}),
\mathbf{V}_2=\mathbf{W}^{\mathcal{V}}_2\mathcal{F}^{q},
\end{equation}
where $\mathbf{W}^{\mathcal{Q}}_{2}$, $\mathbf{W}^{\mathcal{K}}_{2}$, $\mathbf{W}^{\mathcal{V}}_{2}$ are learnable parameters. Our proposed AFL can be written as:
\begin{equation}\label{AFL}
\widehat{\mathcal{P}} = \mathbf{SA}_2\left(\mathrm{Softmax}\left(\frac{\mathbf{Q}_2\mathbf{K}_2^\mathrm{T}}{\sqrt{d}}\right)\mathbf{V}_2\right),
\end{equation}
where $\mathbf{SA}_2(\cdot)$ denotes another self-attention module different from $\mathbf{SA}_1(\cdot)$, $\widehat{\mathcal{P}}\in \mathbb{R}^{M\times C}$ is called anomaly proxies. $\widehat{\mathcal{P}}$ captures abnormal patterns in $\mathcal{F}^{q}$ by comparing reference residuals $\mathrm{Res}(\mathcal{F}^a, \mathcal{F}^n)$ with the query-normal residuals $\mathrm{Res}(\mathcal{F}^{q},\mathcal{F}^{n})$.  $\widehat{\mathcal{P}}$ has the most distinguishable patch feature information in $\mathcal{F}^{q}$. Therefore, we calculate the mean of similarity between each anomaly proxy $\widehat{\mathcal{P}}_{m}$ and the query patch features $\mathcal{F}^q$ as an anomaly-guided anomaly score map $\mathcal{S}_{a}$. Each patch anomaly score is
\begin{equation}
\mathcal{S}^{i}_a = \frac{1}{M}\sum^{M}_{m=1}1-\mathbf{d}(f^{q}_{i},\widehat{\mathcal{P}}_{m}),
\end{equation}
where $\mathcal{S}_{a}\in \mathbb{R}^{L}$. Finally, we merge the initial normal-guided score map and abnormal-guided score map to acquire final anomaly localization results:
\begin{equation}\label{final score}
\mathcal{S} = \mathcal{S}_n + \mathcal{S}_a.
\end{equation}
Following \cite{damm2024anomalydino}, we calculate the image-level anomaly score by averaging the top $1\%$ highest values in the score map $\mathcal{S}\in\mathbb{R}^{L}$. Specifically, we denote this operation as $s=\mathcal{T}_{0.01}(\mathcal{S})$, where $\mathcal{T}_{0.01}(\mathcal{S})$ represents the average of the $1\%$ highest scores in $\mathcal{S}$.

\subsection{Training on Original Domain}
The whole NAGL framework is trained on original domain data. Our optimization objective is that the predicted map $\mathcal{S}$ should be consistent with the query ground truth mask $\mathcal{M}^q$. Therefore, we apply Focal loss \cite{lin2017focal} and Dice loss \cite{li2020dice} to achieve anomaly segmentation training, \ie,
\begin{equation}
\mathcal{L}_{\mathrm{seg}} = \mathbf{Focal}(\mathcal{S}, \mathcal{M}^q) + \mathbf{Dice}(\mathcal{S}, \mathcal{M}^q), 
\end{equation}
where $\mathcal{M}^q$ denotes the downsampled and reshaped ground truth mask of the query image, $\mathbf{Focal}(\cdot)$ and $\mathbf{Dice}(\cdot)$ represent the Focal loss and Dice loss, respectively. Moreover, we also guarantee that the image-level predicted label is consistent with the classification label of the query image. Binary Cross-Entropy (BCE) loss is leveraged to optimize anomaly classification,
\begin{equation}
\mathcal{L}_{\mathrm{cls}} = \mathbf{BCE}(s, y^q), 
\end{equation}
where $s$ denotes predicted image-level anomaly score, $y^q$ is ground truth classification label, and $\mathbf{BCE}(\cdot)$ denotes BCE loss function. Finally, a hyper-parameter $\lambda$ is used to balance the classification and segmentation losses,
\begin{equation}
\mathcal{L} = \mathcal{L}_{\mathrm{cls}}+\lambda\mathcal{L}_{\mathrm{seg}}. 
\end{equation}

\subsection{Inference on Target Domain}
During inference, for given target domain data $D_{target}$, similar to the training phase, we first use the NN method to obtain a normal-guided score map $\mathcal{S}_{n}$. Next, we apply RM and AFL modules to generate an abnormal-guided score map $\mathcal{S}_{a}$. According to Eq.~\eqref{final score}, we merge the two score maps as the final score map $\mathcal{S}\in \mathbb{R}^{L}$. Next, the size of $\mathcal{S}$ is reshaped to $\mathbb{R}^{h\times w}$, and up-sampled to $\mathbb{R}^{H\times W}$. 

\begin{table}[tp]
    \caption{Comparison of the proposed method with the previous methods on MVTecAD and VisA datasets. $N^i$ and $A^i$ represent the number of normal and abnormal reference samples in $i$-shot learning, respectively. Results marked with ${\dag}$ are quoted from \cite{jeong2023winclip}, while those marked with ${\ast}$ are based on our re-implementation. The best/runner-up results are highlighted in \textbf{bold}/\underline{underline}.} 
    \label{tab:exp_industrial}%
    \scriptsize
    \setlength\tabcolsep{3.0pt}
    \centering
      \begin{tabular}{cc|cccccc|cccccc}
      \toprule
      \multirow{3}[4]{*}{Setting} & \multirow{3}[4]{*}{Method} & \multicolumn{6}{c|}{\textbf{MVTecAD}}                 & \multicolumn{6}{c}{\textbf{VisA}} \\
           &      & \multicolumn{3}{c}{Image-level} & \multicolumn{3}{c|}{Pixel-level} & \multicolumn{3}{c}{Image-level} & \multicolumn{3}{c}{Pixel-level} \\
 &      & AUROC & AP & F1-max & AUROC & PRO  & F1-max & AUROC & AP & F1-max & AUROC & PRO  & F1-max \\
      \midrule
      \multirow{6}[2]{*}{$N^1$} & SPADE$^{\dag}$ \cite{SPADE} & 81.0 & 90.6 & 90.3 & 91.2 & 83.9 & 42.4 & 79.5 & 82.0 & 80.7 & 95.6 & 84.1 & 35.5  \\
           & PaDiM$^{\dag}$ \cite{PaDiM} & 76.6 & 88.1 & 88.2 & 89.3 & 73.3 & 40.2 & 88.2 & 62.8 & 75.3 & 89.9 & 64.3 & 17.4  \\
           & PatchCore$^{\dag}$ \cite{Roth_2022_CVPR} & 83.4 & 92.2 & 90.5 & 92.0 & 79.7 & 50.4 & 79.9 & 82.8 & 81.7 & 95.4 & 80.5 & 38.0  \\
           & WinCLIP$^{\dag}$ \cite{jeong2023winclip} & 93.1 & 96.5 & \underline{93.7} & 95.2 & 87.1 & \underline{55.9} & 83.8 & 85.1 & \underline{83.1} & 96.4 & \underline{85.1} & \underline{41.3}  \\
           & PromptAD \cite{li2024promptad} & \underline{94.6} & \underline{97.1} & - & \underline{95.9} & \underline{87.9} & - & \underline{86.9} & \underline{88.4} & - & \underline{96.7} & \underline{85.1} & - \\
           & ResAD$^{\ast}$ \cite{yao2024resad} & 84.8 & 92.7 & 91.2 & 93.4 & 83.3 & 48.2 & 80.9 & 83.7 & 81.3 & 95.9 & 79.6 & 37.8  \\
      \midrule
      $N^1+A^1$ & \textbf{Ours} & \textbf{95.8} & \textbf{97.5} & \textbf{95.7} & \textbf{96.6} & \textbf{92.9} & \textbf{58.9} & \textbf{88.5} & \textbf{89.4} & \textbf{85.4} & \textbf{97.5} & \textbf{91.1} & \textbf{41.8} \\
      \midrule
      \midrule
      \multirow{7}[2]{*}{$N^2$} & SPADE$^{\dag}$ \cite{SPADE} & 82.9 & 91.7 & 91.1 & 92.0 & 85.7 & 44.5 & 80.7 & 82.3 & 81.7 & 96.2 & 85.7 & 40.5  \\
           & PaDiM$^{\dag}$ \cite{PaDiM} & 78.9 & 89.3 & 89.2 & 91.3 & 78.2 & 43.7 & 67.4 & 71.6 & 75.7 & 92.0 & 70.1 & 21.1  \\
           & PatchCore$^{\dag}$ \cite{Roth_2022_CVPR} & 86.3 & 93.8 & 92.0 & 93.3 & 82.3 & 53.0 & 81.6 & 84.8 & 82.5 & 96.1 & 82.6 & 41.0  \\
           & WinCLIP$^{\dag}$ \cite{jeong2023winclip} & 94.4 & 97.0 & \underline{94.4} & 96.0 & 88.4 & \underline{58.4} & 84.6 & 85.8 & 83.0 & 96.8 & \underline{86.2} & \textbf{43.5}  \\
           & PromptAD \cite{li2024promptad} & \underline{95.7} & \underline{97.9} & - & \underline{96.2} & \underline{88.5} & - & \underline{88.3} & \underline{90.0} & - & \underline{97.1} & 85.8 & - \\
           & InCTRL \cite{zhu2024toward} & 94.0 & 96.9 & - & -    & - & -    & 85.8 & 87.7 & - & -    & - & - \\
           & ResAD$^{\ast}$ \cite{yao2024resad} & 87.2 & 93.9 & 92.2 & 94.8 & 85.5 & 50.2 & 86.6 & 88.3 & \underline{84.1} & 96.5 & 82.3 & 40.0  \\
      \midrule
      $N^2+A^1$ & \textbf{Ours} & \textbf{96.8} & \textbf{97.9} & \textbf{96.3} & \textbf{96.8} & \textbf{93.2} & \textbf{59.8} & \textbf{89.8} & \textbf{90.6} & \textbf{86.8} & \textbf{97.6} & \textbf{91.4} & \underline{43.3}  \\
      \midrule
      \midrule
      \multirow{7}[2]{*}{$N^4$} & SPADE$^{\dag}$ \cite{SPADE} & 84.8 & 92.5 & 91.5 & 92.7 & 87.0 & 46.2 & 81.7 & 83.4 & 82.1 & 96.6 & 87.3 & 43.6  \\
           & PaDiM$^{\dag}$ \cite{PaDiM} & 80.4 & 90.5 & 90.2 & 92.6 & 81.3 & 46.1 & 72.8 & 75.6 & 78.0 & 93.2 & 72.6 & 24.6  \\
           & PatchCore$^{\dag}$ \cite{Roth_2022_CVPR} & 88.8 & 94.5 & 92.6 & 94.3 & 84.3 & 55.0 & 85.3 & 87.5 & 84.3 & 96.8 & 84.9 & 43.9  \\
           & WinCLIP$^{\dag}$ \cite{jeong2023winclip} & 95.2 & 97.3 & \underline{94.7} & 96.2 & 89.0 & \underline{59.5} & 87.3 & 88.8 & 84.2 & 97.2 & \underline{87.6} & \textbf{47.0}  \\
           & PromptAD \cite{li2024promptad} & \underline{96.6} & \underline{98.5} & - & \underline{96.5} & \underline{90.5} & - & 89.1 & 90.8 & - & \underline{97.4} & 86.2 & - \\
           & InCTRL \cite{zhu2024toward} & 94.5 & 97.2 & - & -    & - & -    & 87.7 & 90.2 & - & -    & - & - \\
           & ResAD$^{\ast}$ \cite{yao2024resad} & 90.7 & 95.7 & 93.9 & 95.8 & 88.7 & 53.0 & \underline{89.3} & \underline{90.7} & \underline{86.5} & 96.8 & 84.1 & 41.6  \\
      \midrule
      $N^4+A^1$ & \textbf{Ours} & \textbf{97.1} & \textbf{98.0} & \textbf{96.4} & \textbf{97.0} & \textbf{93.5} & \textbf{60.1} & \textbf{91.2} & \textbf{91.7} & \textbf{87.8} & \textbf{97.8} & \textbf{91.5} & \underline{44.0} \\
      \bottomrule
      \end{tabular}%
  \end{table}%

\section{Experiment}

\subsection{Experimental Setup}
\label{sec:set_up}

\paragraph{Datasets.} To validate the efficiency of our NAGL framework, we construct three benchmarks using the MVTecAD \cite{MVTec}, VisA \cite{VisA}, and BraTS \cite{menze2014brats} datasets. The benchmarks include (1) training on VisA and testing on MVTecAD, (2) training on MVTecAD and testing on VisA, and (3) training on MVTecAD and testing on BraTS. The first two benchmarks assess generalization capabilities across different industrial domains, while the third evaluates cross-domain transfer capabilities from industrial to medical applications. During inference, the normal references are sampled from the training set of target datasets, while abnormal references are sampled from each anomaly type. Further details are provided in Appendix~\ref{sec:supp_a}.

\paragraph{Evaluation Metrics.} We evaluate both image-level anomaly classification and pixel-level segmentation performance using three metrics for each task, following previous works \cite{jeong2023winclip,damm2024anomalydino,yao2024resad}. For detection performance, we employ the Area Under the Receiver-Operator Curve (AUROC, \cite{PaDiM}), the maximum F1-score at the optimal threshold (F1-max), and the Average Precision (AP), calculated using image-level anomaly scores. Similarly, for segmentation performance, we utilize the AUROC, F1-max, and per-region overlap (PRO) \cite{bergmann2020aupro,MVTec} metrics, computed using pixel-wise anomaly scores.

\paragraph{Implementation Details.} Our approach employs the ViT-based \cite{oquab2024dinov} model as the vision encoder, specifically utilizing its most lightweight version (ViT-S, $21M$ parameters) to ensure low latency in practical applications. We freeze the parameters of the vision encoder throughout the experiments and only update the parameters of the proposed attention modules. The model is optimized using AdamW \cite{loshchilov2018decoupled} with an initial learning rate of $1\times10^{-5}$, which is reduced by a factor of $0.1$ at epoch $10$ and $15$. The training process converges within $20$ epochs, with each epoch comprising $500$ sampled episodes. Input images are resized to $448 \times 448$ resolution without data augmentation. We set the number of learnable proxies ($\mathcal{P}$) to $M=25$ by default and use a loss balance weight ($\lambda$) of $1.0$. Following previous works \cite{jeong2023winclip,li2024promptad}, we set the number of normal references as $K_1\in[1,2,4]$. Considering the scarcity of abnormal samples, we only use one abnormal reference ($K_2=1$), making our approach highly applicable in real-world scenarios. The implementation is based on PyTorch$@2.1.1$, and the experiments are conducted on a single NVIDIA RTX 4090 24GB GPU. To ensure statistical reliability, we report results averaged across $3$ independent runs with different random seeds.

\paragraph{Comparison Methods.} We select some representative few-normal-shot AD methods to comparison, including SPADE \cite{SPADE}, PaDiM \cite{PaDiM}, PatchCore \cite{Roth_2022_CVPR}, RegAD \cite{huang2022registration}, WinCLIP \cite{jeong2023winclip}, MVFA \cite{huang2024adapting}, and PromptAD \cite{li2024promptad}. For generalist AD, we primarily compared our approach with InCTRL \cite{zhu2024toward} and ResAD \cite{yao2024resad}. Tab.~\ref{tab:exp_industrial} and Tab.~\ref{tab:exp_medical} present the comparative results of our method against these baselines. The results marked with ${\dag}$ are reported by \cite{jeong2023winclip}. Due to incomplete ResAD results (only reporting 2/4-shot settings with AUROC metrics, we reproduce the results using their official code (marked with ${\ast}$). For further details, please refer to Appendix~\ref{sec:supp_d}.

\begin{table}[tp] 
\scriptsize
\begin{minipage}[t]{0.46\textwidth} 
    \caption{The image/pixel-level AUROC scores of the proposed method and previous methods on the BraTS dataset. Results marked with ${\ast}$ are based on our re-implementation. Other results are reported from \cite{yao2024resad}.}
    \label{tab:exp_medical}%
    \setlength\tabcolsep{5pt}
    \renewcommand\arraystretch{1.06}
    \centering
    \begin{tabular}{cc|ccc}
       \toprule
       Setting & Method & Image & Pixel & Mean \\
       \midrule
       $N^1$ & ResAD$^{\ast}$ \cite{yao2024resad} & \underline{73.5}  & \underline{91.0}  & \underline{82.3}  \\
       \midrule
       $N^1+A^1$ & \textbf{Ours}  & \textbf{78.1}  & \textbf{96.5}  & \textbf{87.3}  \\
       \midrule
       \midrule
       \multirow{7}[2]{*}{$N^2$} & SPADE \cite{SPADE} & 58.0  & 92.8  & 75.4  \\
             & PaDiM \cite{PaDiM} & 59.4  & 90.2  & 74.8  \\
             & PatchCore \cite{Roth_2022_CVPR} & 58.2  & \underline{93.5}  & 75.9  \\
             & RegAD \cite{huang2022registration} & 54.6  & 81.4  & 68.0  \\
             & WinCLIP \cite{jeong2023winclip} & 55.9  & 91.5  & 73.7  \\
             & InCTRL \cite{zhu2024toward} & \underline{74.6}  & -     & -  \\
             & ResAD$^{\ast}$ \cite{yao2024resad} & 66.2  & 91.5  & \underline{78.9}  \\
       \midrule
       $N^2+A^1$ & \textbf{Ours}  & \textbf{82.1}  & \textbf{96.8}  & \textbf{89.5}  \\
       \midrule
       \midrule
       \multirow{7}[2]{*}{$N^4$} & SPADE \cite{SPADE} & 66.3  & 94.8  & 80.6  \\
             & PaDiM \cite{PaDiM} & 60.6  & 94.5  & 77.6  \\
             & PatchCore \cite{Roth_2022_CVPR} & 71.2  & \underline{95.9}  & 83.6  \\
             & RegAD \cite{huang2022registration} & 60.0  & 87.3  & 73.7  \\
             & WinCLIP \cite{jeong2023winclip} & 67.3  & 93.2  & 80.3  \\
             & MVFA$^{\ast}$ \cite{huang2024adapting} & 75.2  & 92.7     & 84.0  \\
             & InCTRL \cite{zhu2024toward} & \underline{76.9}  & -     & -  \\
             & ResAD$^{\ast}$ \cite{yao2024resad} & 74.9  & 94.3  & \underline{84.6}  \\
       \midrule
       $N^4+A^1$ & \textbf{Ours}  & \textbf{84.9}  & \textbf{97.1}  & \textbf{91.0}  \\
       \bottomrule
    \end{tabular}%
\end{minipage}
\hfill 
\begin{minipage}[t]{0.52\textwidth} 
    \caption{Ablation study on the MVTecAD and VisA datasets. Image/pixel-level AUROC are reported. $N^1$/$A^1$ represents the one normal/abnormal reference sample, "NAGL" indicates the use of our proposed framework. There are four different settings. {\romannumeral1}/{\romannumeral2} uses only normal/abnormal samples through NN search, {\romannumeral3} merges the results of {\romannumeral1} and {\romannumeral2} without additional processing, and {\romannumeral4} includes our proposed process.} 
    \label{tab:exp_ablation}%
    \renewcommand\arraystretch{0.8}
    \setlength\tabcolsep{3.9pt}
    \centering
    \begin{tabular}{ccccccccc}
    \toprule
          & \multirow{2}[2]{*}{$N^1$} & \multirow{2}[2]{*}{$A^1$} & \multirow{2}[2]{*}{NAGL} & \multicolumn{2}{c}{MVTecAD} & \multicolumn{2}{c}{VisA} & \multirow{2}[2]{*}{Mean} \\
          &       &       &       & Image & Pixel & Image & Pixel &  \\
    \midrule
    \romannumeral1 & \checkmark &       &       & 93.2  & 94.5  & 81.5  & 95.3  & 91.1  \\
    \romannumeral2 &  & \checkmark &       & 70.1  & 83.3  & 58.8  & 84.5  & 74.2  \\
    \romannumeral3 & \checkmark & \checkmark &       & 90.7  & 92.1  & 77.2  & 93.5  & 88.4  \\
    \romannumeral4 & \checkmark & \checkmark & \checkmark & \textbf{95.8} & \textbf{96.6} & \textbf{88.5} & \textbf{97.5} & \textbf{94.6} \\
    \bottomrule
    \end{tabular}%
    \vspace{\floatsep} 
    \caption{Comparison of the total parameters, training time, and inference speed of the proposed method with InCTRL and ResAD. The training time of InCTRL and ResAD is measured under the $N^1$ setting, while our method is based on the $N^1+A^1$ setting.}
    \label{tab:exp_efficiency}%
    \renewcommand\arraystretch{0.8}
    \centering
    \begin{tabular}{c|ccc}
       \toprule
       \multirow{2}[2]{*}{Method} & Total & Training & Inference \\
             & Parameters (M) & Time (H) & Speed (FPS) \\
       \midrule
       InCTRL & 117.5  & 0.7     & 1.2  \\
       ResAD & 59.2  & 20.6 & 7.8  \\
       Ours  & \textbf{24.4}  & \textbf{0.3}  & \textbf{17.1}  \\
       \bottomrule
    \end{tabular}%
\end{minipage}
\end{table} 

\subsection{Main Results}

Tab.~\ref{tab:exp_industrial} and Tab.~\ref{tab:exp_medical} present comparative results between our proposed method and existing approaches across MVTecAD, VisA, and BraTS datasets. The experimental results demonstrate three key advantages of our method. Additionally, to demonstrate the generalization capabilities, we also present the performance on MVTec3D \cite{bergmann2021mvtec}, MVTecLOCO \cite{bergmann2022beyond}, BTAD \cite{mishra2021btad}, and MPDD \cite{jezek2021mpdd} datasets in Appendix~\ref{sec:supp_g}.

\paragraph{Benefits of Single Abnormal Sample.} Incorporating just one abnormal reference sample ($A^1$) yields substantial performance gains across all scenarios. On industrial datasets (Tab. \ref{tab:exp_industrial}), our $N^1+A^1$ setting achieves $95.8\%$ and $96.6\%$ for image and pixel AUROC, respectively on MVTecAD, surpassing the best $N^4$ baseline method (WinCLIP: $95.2\%$, $96.2\%$) by $\textbf{0.6}$ and $\textbf{0.4}$ percentage points while using fewer reference samples. A similar trend is observed on VisA, where our $N^2+A^1$ setting achieves $89.8\%$ and $97.6\%$ for image and pixel AUROC, outperforming the best baseline method (ResAD with $N^4$: $89.3\%$, $96.8\%$) by $\textbf{0.5}$ and $\textbf{0.8}$ percentage points, respectively. These results show significant benefits from abnormal reference data.

\paragraph{Generalization.} Our method exhibits superior generalization capabilities compared to ResAD across multiple datasets. On MVTecAD, under the $N^1+A^1$ setting, our approach achieves $95.7\%$ image F1-max and $58.9\%$ pixel F1-max, outperforming ResAD by $\textbf{4.5}$ and $\textbf{10.7}$ percentage points, respectively. Similarly, on VisA, we attain $85.4\%$ image F1-max and $41.8\%$ pixel F1-max, exceeding ResAD by $\textbf{4.1}$ and $\textbf{4.0}$ percentage points. Furthermore, as shown in Tab.~\ref{tab:exp_medical}, our method demonstrates superior cross-domain generalization, achieving $89.5\%$ average AUROC on the BraTS medical dataset with just two normal and one abnormal sample, a $\textbf{10.6}$ percentage point improvement over ResAD ($78.9\%$ with $N^2$ setting). Notably, across all other experimental settings, our method consistently outperforms existing approaches in both anomaly detection and segmentation. These results clearly demonstrate the effectiveness of our method in generalist AD.

\paragraph{Scalability.} Through comparative analysis, our method demonstrates excellent scalability as the number of normal samples increases across all three datasets. For example, on VisA, we observe consistent performance improvements when progressing from $N^1+A^1$ to $N^2+A^1$ and $N^4+A^1$ settings. Specifically, the image F1-max score increases from $85.4\%$ to $86.8\%$ and $87.8\%$, while the pixel F1-max score improves from $41.8\%$ to $43.3\%$ and reaches $44.0\%$. On BraTS, the average AUROC improves from $87.3\%$ to $89.5\%$ and reaches $91.0\%$. These results indicate that our method effectively utilizes additional normal samples to enhance the detection ability.

\begin{figure}[tp]
    \centering
    \includegraphics[width=1.0\linewidth]{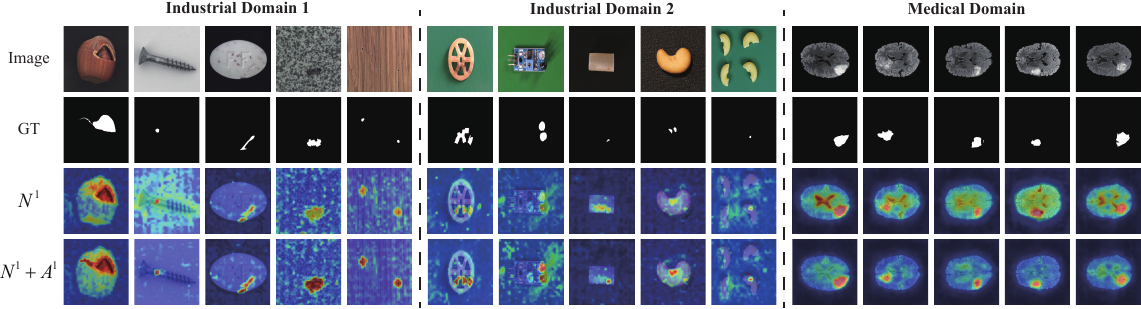}
    \caption{Qualitative results. The first row displays the input images, and the second row shows the ground truth. The third row illustrates anomaly score maps using $1$ normal sample, while the bottom row shows anomaly score maps produced by our method using $1$ normal and $1$ abnormal samples.}
    \label{fig:exp_qualitative}
\end{figure}

\subsection{Ablation Study}
To further investigate the effectiveness of our method, we conduct an ablation study on the MVTecAD and VisA datasets. Tab.~\ref{tab:exp_ablation} presents the results from different experimental settings. The results clearly demonstrate several important findings.  

First, comparing the results of {\romannumeral1}, {\romannumeral2}, and {\romannumeral3}, we observe that naively incorporating abnormal references leads to significant performance degradation both at the image and pixel levels. This aligns with our argument in Sec.~\ref{sec:intro} that the abnormal features cause severe false activation, where the unrefined abnormal features introduce misleading guidance that incorrectly highlights normal patterns as anomalies, thereby compromising detection accuracy.

Second, when comparing settings {\romannumeral3} and {\romannumeral4}, the benefits of our proposed method become evident. By implementing the NAGL framework in setting {\romannumeral4}, our method manages to effectively mitigate the false activation problem observed when naively using an abnormal reference. Properly mining the abnormal variations in residual space to guide detection can effectively suppress the over-activation, resulting in $6.2$ average AUROC improvements from $88.4\%$ to $94.6\%$. 

Finally, the comparison between {\romannumeral1} and {\romannumeral4} reveals that our method fully unleashes the potential of abnormal references, with gains of $+2.6$ Image AUROC on MVTecAD and $+7.0$ on VisA, proving that properly processed abnormal samples provide complementary discriminative signals. 

\subsection{Analysis of two attentions in RM-AFL module} 
In the first attention (RM), we set the value as $\text{Res}(\mathcal{F}^a,\mathcal{F}^n)$, which provides the \textit{normal-abnormal differences} in the residual space. We define the key as $\mathcal{F}^a$, which are the original abnormal patterns. Since the query is learnable, the output is the optimally aggregated residual features (termed as residual proxies $\tilde{\mathcal{P}}$).

The second attention (AFL) aims to learn query-related abnormal patterns by comparing \textit{normal-abnormal differences} and \textit{normal-query differences}. We set the query as $\tilde{\mathcal{P}}$ and the key as $\text{Res}(\mathcal{F}^q,\mathcal{F}^n)$. This \textit{comparison} is achieved by computing the attention map between $\tilde{\mathcal{P}}$ and $\text{Res}(\mathcal{F}^q,\mathcal{F}^n)$. Then the query features $\mathcal{F}^q$ are aggregated by the attention map to obtain the query-related abnormal patterns (termed as anomaly proxies $\hat{\mathcal{P}}$).

The residual proxies $\tilde{\mathcal{P}}$ represent abnormal patterns of reference abnormal samples in residual space, and anomaly proxies $\hat{\mathcal{P}}$ represent abnormal patterns of the query sample in vision space. As validated in Appendix~\ref{sec:supp_b}, residual features exhibit a common distribution, even among residual features of different anomalies. Residual features of known anomalies can provide references; if the residual feature of a query is similar to them, the corresponding visual feature is likely to be an anomaly.

\subsection{Efficiency Analysis}
\label{sec:eff}
We compare the efficiency of our method with InCTRL and ResAD in terms of total parameters, training time, and inference speed. Tab.~\ref{tab:exp_efficiency} shows that our method is more efficient than InCTRL and ResAD regarding total parameters and training time. Our method has $24.4$ million parameters, which is around $\textbf{5}\times$ \textbf{smaller} than InCTRL. Our method only requires $0.3$ hours for training, which is $\textbf{69}\times$ \textbf{faster} than ResAD. In terms of inference speed, our method achieves $17.1$ FPS, which is $\textbf{14}\times$ \textbf{faster} than InCTRL and $\textbf{2}\times$ \textbf{faster} than ResAD. These results demonstrate the efficiency of our method for real-world deployment. 

\subsection{Qualitative Results}
Fig.~\ref{fig:exp_qualitative} presents qualitative results from both industrial and medical datasets. Comparing the score maps, we observe that our method effectively generates more accurate results, which demonstrate that incorporating abnormal reference samples provides valuable guidance boundaries for the detection process, and our method effectively leverages this information to improve detection accuracy.

\section{Conclusion}
\label{sec:conclusion}
In this paper, we introduce a novel task for anomaly detection using both normal and abnormal references. Our approach addresses the limitations of traditional methods by leveraging limited known anomalies to guide the detection of unseen anomalies. We propose a NAGL framework that extracts discriminative features in the residual space, effectively capturing the essence of anomalies. Experimental results across multiple benchmarks demonstrate that our method significantly outperforms existing approaches, particularly in detecting challenging unseen anomalies. The performance gains are consistent across various datasets and settings, highlighting the robustness of our approach. This work opens new research directions for anomaly detection with limited supervision, with potential applications in industrial inspection and medical diagnosis. Future research could delve into more effective methods for leveraging scarce anomaly samples or expanding prompts. For instance, it could involve enabling language-based descriptions for normal or abnormal references. 

\section{Acknowledgements}
This work is supported by the National Natural Science Foundation of China (No. 62471405, 62331003, 62301451), Jiangsu Basic Research Program Natural Science Foundation (BK20241814), Suzhou Basic Research Program (SYG202316) and XJTLU REF-22-01-010, XJTLU AI University Research Center, Jiangsu Province Engineering Research Center of Data Science and Cognitive Computation at XJTLU and SIP AI innovation platform (YZCXPT2022103).

{
    \small
    \bibliographystyle{plain}
    \bibliography{main}

\begin{thebibliography}{10}

\bibitem{bergmann2022beyond}
Paul Bergmann, Kilian Batzner, Michael Fauser, David Sattlegger, and Carsten Steger.
\newblock Beyond dents and scratches: Logical constraints in unsupervised anomaly detection and localization.
\newblock {\em International Journal of Computer Vision}, 130(4):947--969, 2022.

\bibitem{MVTec}
Paul Bergmann, Michael Fauser, David Sattlegger, and Carsten Steger.
\newblock {MVTec AD} -- a comprehensive real-world dataset for unsupervised anomaly detection.
\newblock In {\em CVPR}, 2019.

\bibitem{bergmann2020aupro}
Paul Bergmann, Michael Fauser, David Sattlegger, and Carsten Steger.
\newblock Uninformed students: Student-teacher anomaly detection with discriminative latent embeddings.
\newblock In {\em CVPR}, 2020.

\bibitem{bergmann2021mvtec}
Paul Bergmann, Xin Jin, David Sattlegger, and Carsten Steger.
\newblock The mvtec 3d-ad dataset for unsupervised 3d anomaly detection and localization.
\newblock {\em arXiv preprint arXiv:2112.09045}, 2021.

\bibitem{cao2023anomaly}
Tri Cao, Jiawen Zhu, and Guansong Pang.
\newblock Anomaly detection under distribution shift.
\newblock In {\em ICCV}, 2023.

\bibitem{10884560}
Yunkang Cao, Xiaohao Xu, Yuqi Cheng, Chen Sun, Zongwei Du, Liang Gao, and Weiming Shen.
\newblock Personalizing vision-language models with hybrid prompts for zero-shot anomaly detection.
\newblock {\em IEEE Transactions on Cybernetics}, 55(4):1917--1929, 2025.

\bibitem{cao2024adaclip}
Yunkang Cao, Jiangning Zhang, Luca Frittoli, Yuqi Cheng, Weiming Shen, and Giacomo Boracchi.
\newblock {AdaCLIP}: Adapting {CLIP} with hybrid learnable prompts for zero-shot anomaly detection.
\newblock In {\em ECCV}, 2024.

\bibitem{chen2022deep}
Yuanhong Chen, Yu~Tian, Guansong Pang, and Gustavo Carneiro.
\newblock Deep one-class classification via interpolated gaussian descriptor.
\newblock In {\em AAAI}, 2022.

\bibitem{chen2024filter}
Zining Chen, Xingshuang Luo, Weiqiu Wang, Zhicheng Zhao, Fei Su, and Aidong Men.
\newblock Filter or compensate: Towards invariant representation from distribution shift for anomaly detection.
\newblock In {\em AAAI}, 2025.

\bibitem{SPADE}
Niv Cohen and Yedid Hoshen.
\newblock Sub-image anomaly detection with deep pyramid correspondences.
\newblock {\em ArXiv}, abs/2005.02357, 2020.

\bibitem{costanzino2024multimodal}
Alex Costanzino, Pierluigi~Zama Ramirez, Giuseppe Lisanti, and Luigi Di~Stefano.
\newblock Multimodal industrial anomaly detection by crossmodal feature mapping.
\newblock In {\em CVPR}, 2024.

\bibitem{damm2024anomalydino}
Simon Damm, Mike Laszkiewicz, Johannes Lederer, and Asja Fischer.
\newblock {AnomalyDINO}: Boosting patch-based few-shot anomaly detection with dinov2.
\newblock {\em arXiv}, 2024.

\bibitem{PaDiM}
Thomas Defard, Aleksandr Setkov, Angelique Loesch, and Romaric Audigier.
\newblock {PaDiM}: A patch distribution modeling framework for anomaly detection and localization.
\newblock In {\em ICPR}, 2021.

\bibitem{ding2022catching}
Choubo Ding, Guansong Pang, and Chunhua Shen.
\newblock Catching both gray and black swans: Open-set supervised anomaly detection.
\newblock In {\em CVPR}, 2022.

\bibitem{fang2023fastrecon}
Zheng Fang, Xiaoyang Wang, Haocheng Li, Jiejie Liu, Qiugui Hu, and Jimin Xiao.
\newblock {FastRecon}: Few-shot industrial anomaly detection via fast feature reconstruction.
\newblock In {\em ICCV}, 2023.

\bibitem{gu2024rethinking}
Zhihao Gu, Jiangning Zhang, Liang Liu, Xu~Chen, Jinlong Peng, Zhenye Gan, Guannan Jiang, Annan Shu, Yabiao Wang, and Lizhuang Ma.
\newblock Rethinking reverse distillation for multi-modal anomaly detection.
\newblock In {\em AAAI}, 2024.

\bibitem{guo2023recontrast}
Jia Guo, Shuai Lu, Lize Jia, Weihang Zhang, and Huiqi Li.
\newblock Recontrast: Domain-specific anomaly detection via contrastive reconstruction.
\newblock In {\em NeurIPS}, volume~36, pages 10721--10740, 2023.

\bibitem{guo2025dinomaly}
Jia Guo, Shuai Lu, Weihang Zhang, Fang Chen, Huiqi Li, and Hongen Liao.
\newblock Dinomaly: The less is more philosophy in multi-class unsupervised anomaly detection.
\newblock In {\em CVPR}, pages 20405--20415, 2025.

\bibitem{he2024diffusion}
Haoyang He, Jiangning Zhang, Hongxu Chen, Xuhai Chen, Zhishan Li, Xu~Chen, Yabiao Wang, Chengjie Wang, and Lei Xie.
\newblock A diffusion-based framework for multi-class anomaly detection.
\newblock In {\em AAAI}, 2024.

\bibitem{he2023transformers}
Kelei He, Chen Gan, Zhuoyuan Li, Islem Rekik, Zihao Yin, Wen Ji, Yang Gao, Qian Wang, Junfeng Zhang, and Dinggang Shen.
\newblock Transformers in medical image analysis.
\newblock {\em Intelligent Medicine}, 3(1):59--78, 2023.

\bibitem{huang2022registration}
Chaoqin Huang, Haoyan Guan, Aofan Jiang, Ya~Zhang, Michael Spratling, and Yan-Feng Wang.
\newblock Registration based few-shot anomaly detection.
\newblock In {\em ECCV}, pages 303--319, 2022.

\bibitem{huang2024adapting}
Chaoqin Huang, Aofan Jiang, Jinghao Feng, Ya~Zhang, Xinchao Wang, and Yanfeng Wang.
\newblock Adapting visual-language models for generalizable anomaly detection in medical images.
\newblock In {\em CVPR}, pages 11375--11385, 2024.

\bibitem{jeong2023winclip}
Jongheon Jeong, Yang Zou, Taewan Kim, Dongqing Zhang, Avinash Ravichandran, and Onkar Dabeer.
\newblock {WinCLIP}: Zero-/few-shot anomaly classification and segmentation.
\newblock In {\em CVPR}, 2023.

\bibitem{jezek2021mpdd}
Stepan Jezek, Martin Jonak, Radim Burget, Pavel Dvorak, and Milos Skotak.
\newblock Deep learning-based defect detection of metal parts: Evaluating current methods in complex conditions.
\newblock In {\em ICUMT}, 2021.

\bibitem{jiang2025mmad}
Xi~Jiang, Jian Li, Hanqiu Deng, Yong Liu, Bin-Bin Gao, Yifeng Zhou, Jialin Li, Chengjie Wang, and Feng Zheng.
\newblock {MMAD}: A comprehensive benchmark for multimodal large language models in industrial anomaly detection.
\newblock In {\em ICLR}, 2025.

\bibitem{jiang2022softpatch}
Xi~Jiang, Jianlin Liu, Jinbao Wang, Qiang Nie, Kai Wu, Yong Liu, Chengjie Wang, and Feng Zheng.
\newblock {SoftPatch}: Unsupervised anomaly detection with noisy data.
\newblock In {\em NeurIPS}, 2022.

\bibitem{krizhevsky2012imagenet}
Alex Krizhevsky, Ilya Sutskever, and Geoffrey~E Hinton.
\newblock Imagenet classification with deep convolutional neural networks.
\newblock In {\em NeurIPS}, volume~25, 2012.

\bibitem{lee2024text}
Mingyu Lee and Jongwon Choi.
\newblock Text-guided variational image generation for industrial anomaly detection and segmentation.
\newblock In {\em CVPR}, 2024.

\bibitem{lei2023pyramidflow}
Jiarui Lei, Xiaobo Hu, Yue Wang, and Dong Liu.
\newblock {PyramidFlow}: High-resolution defect contrastive localization using pyramid normalizing flow.
\newblock In {\em CVPR}, 2023.

\bibitem{Li_2021_CVPR}
Chun-Liang Li, Kihyuk Sohn, Jinsung Yoon, and Tomas Pfister.
\newblock {CutPaste}: Self-supervised learning for anomaly detection and localization.
\newblock In {\em CVPR}, 2021.

\bibitem{li2024promptad}
Xiaofan Li, Zhizhong Zhang, Xin Tan, Chengwei Chen, Yanyun Qu, Yuan Xie, and Lizhuang Ma.
\newblock {PromptAD}: Learning prompts with only normal samples for few-shot anomaly detection.
\newblock In {\em CVPR}, 2024.

\bibitem{li2020dice}
Xiaoya Li, Xiaofei Sun, Yuxian Meng, Junjun Liang, Fei Wu, and Jiwei Li.
\newblock Dice loss for data-imbalanced nlp tasks.
\newblock In {\em ACL}, 2020.

\bibitem{li2024onetonormal}
Yiyue Li, Shaoting Zhang, Kang Li, and Qicheng Lao.
\newblock {One-to-Normal}: Anomaly personalization for few-shot anomaly detection.
\newblock In {\em NeurIPS}, 2024.

\bibitem{lin2024comprehensive}
Jiang Lin and Yaping Yan.
\newblock A comprehensive augmentation framework for anomaly detection.
\newblock In {\em AAAI}, 2024.

\bibitem{lin2017focal}
Tsung-Yi Lin, Priya Goyal, Ross Girshick, Kaiming He, and Piotr Doll{\'a}r.
\newblock Focal loss for dense object detection.
\newblock In {\em ICCV}, 2017.

\bibitem{liu2024real3d}
Jiaqi Liu, Guoyang Xie, Ruitao Chen, Xinpeng Li, Jinbao Wang, Yong Liu, Chengjie Wang, and Feng Zheng.
\newblock {Real3D-AD}: A dataset of point cloud anomaly detection.
\newblock In {\em NeurIPS}, 2024.

\bibitem{liu2023diversity}
Wenrui Liu, Hong Chang, Bingpeng Ma, Shiguang Shan, and Xilin Chen.
\newblock Diversity-measurable anomaly detection.
\newblock In {\em CVPR}, 2023.

\bibitem{long2015fully}
Jonathan Long, Evan Shelhamer, and Trevor Darrell.
\newblock Fully convolutional networks for semantic segmentation.
\newblock In {\em CVPR}, pages 3431--3440, 2015.

\bibitem{loshchilov2018decoupled}
Ilya Loshchilov and Frank Hutter.
\newblock Decoupled weight decay regularization.
\newblock In {\em ICLR}, 2019.

\bibitem{lu2023hierarchical}
Ruiying Lu, YuJie Wu, Long Tian, Dongsheng Wang, Bo~Chen, Xiyang Liu, and Ruimin Hu.
\newblock Hierarchical vector quantized transformer for multi-class unsupervised anomaly detection.
\newblock In {\em NeurIPS}, 2023.

\bibitem{luo2025INP-Former}
Wei Luo, Yunkang Cao, Haiming Yao, Xiaotian Zhang, Jianan Lou, Yuqi Cheng, Weiming Shen, and Wenyong Yu.
\newblock Exploring intrinsic normal prototypes within a single image for universal anomaly detection.
\newblock In {\em CVPR}, pages 9974--9983, June 2025.

\bibitem{lv2025oneforall}
Wenxi Lv, Qinliang Su, and Wenchao Xu.
\newblock One-for-all few-shot anomaly detection via instance-induced prompt learning.
\newblock In {\em ICLR}, 2025.

\bibitem{author2025mvrec}
Shuai Lyu, Fangjian Liao, Zeqi Ma, Rongchen Zhang, Dongmei Mo, and Waikeung Wong.
\newblock {MVREC}: A general few-shot defect classification model using multi-view region-context.
\newblock In {\em AAAI}, 2025.

\bibitem{mcintosh2023inter}
Declan McIntosh and Alexandra~Branzan Albu.
\newblock Inter-realization channels: Unsupervised anomaly detection beyond one-class classification.
\newblock In {\em CVPR}, 2023.

\bibitem{menze2014brats}
Bjoern~H Menze, Andras Jakab, Stefan Bauer, Jayashree Kalpathy-Cramer, Keyvan Farahani, Justin Kirby, Yuliya Burren, Nicole Porz, Johannes Slotboom, Roland Wiest, et~al.
\newblock The multimodal brain tumor image segmentation benchmark (brats).
\newblock {\em IEEE transactions on medical imaging}, 34(10):1993--2024, 2014.

\bibitem{mishra2021btad}
Pankaj Mishra, Riccardo Verk, Daniele Fornasier, Claudio Piciarelli, and Gian~Luca Foresti.
\newblock {VT-ADL}: A vision transformer network for image anomaly detection and localization.
\newblock In {\em ISIE}, pages 01--06, 2021.

\bibitem{oquab2024dinov}
Maxime Oquab, Timoth{\'e}e Darcet, Th{\'e}o Moutakanni, Huy~V. Vo, Marc Szafraniec, Vasil Khalidov, Pierre Fernandez, Daniel HAZIZA, Francisco Massa, Alaaeldin El-Nouby, Mido Assran, Nicolas Ballas, Wojciech Galuba, Russell Howes, Po-Yao Huang, Shang-Wen Li, Ishan Misra, Michael Rabbat, Vasu Sharma, Gabriel Synnaeve, Hu~Xu, Herve Jegou, Julien Mairal, Patrick Labatut, Armand Joulin, and Piotr Bojanowski.
\newblock {DINO}v2: Learning robust visual features without supervision.
\newblock {\em Transactions on Machine Learning Research}, 2024.

\bibitem{Roth_2022_CVPR}
Karsten Roth, Latha Pemula, Joaquin Zepeda, Bernhard Sch\"olkopf, Thomas Brox, and Peter Gehler.
\newblock Towards total recall in industrial anomaly detection.
\newblock In {\em CVPR}, 2022.

\bibitem{strater2024generalad}
Luc~PJ Str{\"a}ter, Mohammadreza Salehi, Efstratios Gavves, Cees~GM Snoek, and Yuki~M Asano.
\newblock Generalad: Anomaly detection across domains by attending to distorted features.
\newblock In {\em ECCV}, 2024.

\bibitem{tao2024kernel}
Fenfang Tao, Guo-Sen Xie, Fang Zhao, and Xiangbo Shu.
\newblock Kernel-aware graph prompt learning for few-shot anomaly detection.
\newblock In {\em AAAI}, 2025.

\bibitem{tien2023revisiting}
Tran~Dinh Tien, Anh~Tuan Nguyen, Nguyen~Hoang Tran, Ta~Duc Huy, Soan Duong, Chanh D~Tr Nguyen, and Steven~QH Truong.
\newblock Revisiting reverse distillation for anomaly detection.
\newblock In {\em CVPR}, 2023.

\bibitem{vaswani2017attention}
Ashish Vaswani, Noam Shazeer, Niki Parmar, Jakob Uszkoreit, Llion Jones, Aidan~N Gomez, {\L}ukasz Kaiser, and Illia Polosukhin.
\newblock Attention is all you need.
\newblock In {\em NeurIPS}, 2017.

\bibitem{wang2024cnc}
Xiaolei Wang, Xiaoyang Wang, Huihui Bai, Eng~Gee Lim, and Jimin Xiao.
\newblock {CNC}: Cross-modal normality constraint for unsupervised multi-class anomaly detection.
\newblock In {\em AAAI}, 2025.

\bibitem{wang2025cnc}
Xiaolei Wang, Xiaoyang Wang, Huihui Bai, Eng~Gee Lim, and Jimin Xiao.
\newblock Cnc: Cross-modal normality constraint for unsupervised multi-class anomaly detection.
\newblock In {\em AAAI}, volume~39, pages 7943--7951, 2025.

\bibitem{wang2025dec}
Xiaolei Wang, Xiaoyang Wang, Huihui Bai, Eng~Gee Lim, and Jimin Xiao.
\newblock Decad: Decoupling anomalies in latent space for multi-class unsupervised anomaly detection.
\newblock In {\em ICCV}, 2025.

\bibitem{wang2025icc}
Xiaolei Wang, Xiaoyang Wang, Huihui Bai, Eng~Gee Lim, and Jimin Xiao.
\newblock Icc: Intra-cluster contraction for pedestrian anomaly detection under one-class classification setting.
\newblock {\em IEEE Transactions on Artificial Intelligence}, 2025.

\bibitem{wang2024towards}
Xiaoyang Wang, Huihui Bai, Limin Yu, Yao Zhao, and Jimin Xiao.
\newblock Towards the uncharted: Density-descending feature perturbation for semi-supervised semantic segmentation.
\newblock In {\em CVPR}, pages 3303--3312, 2024.

\bibitem{xie2023pushing}
Guoyang Xie, Jinbao Wang, Jiaqi Liu, Yaochu Jin, and Feng Zheng.
\newblock Pushing the limits of fewshot anomaly detection in industry vision: {Graphcore}.
\newblock In {\em ICLR}, 2023.

\bibitem{yang2024promptable}
Hui-Yue Yang, Hui Chen, Ao~Wang, Kai Chen, Zijia Lin, Yongliang Tang, Pengcheng Gao, Yuming Quan, Jungong Han, and Guiguang Ding.
\newblock Promptable anomaly segmentation with sam through self-perception tuning.
\newblock In {\em AAAI}, 2025.

\bibitem{yao2024resad}
Xincheng Yao, Zixin Chen, Chao Gao, Guangtao Zhai, and Chongyang Zhang.
\newblock Res{AD}: A simple framework for class generalizable anomaly detection.
\newblock In {\em NeurIPS}, 2024.

\bibitem{yao2023explicit}
Xincheng Yao, Ruoqi Li, Jing Zhang, Jun Sun, and Chongyang Zhang.
\newblock Explicit boundary guided semi-push-pull contrastive learning for supervised anomaly detection.
\newblock In {\em CVPR}, 2023.

\bibitem{zavrtanik2021draem}
Vitjan Zavrtanik, Matej Kristan, and Danijel Sko{\v{c}}aj.
\newblock {DRAEM}-a discriminatively trained reconstruction embedding for surface anomaly detection.
\newblock In {\em ICCV}, 2021.

\bibitem{zhang2023prototypical}
Hui Zhang, Zuxuan Wu, Zheng Wang, Zhineng Chen, and Yu-Gang Jiang.
\newblock Prototypical residual networks for anomaly detection and localization.
\newblock In {\em CVPR}, 2023.

\bibitem{zhang2024realnet}
Ximiao Zhang, Min Xu, and Xiuzhuang Zhou.
\newblock {RealNet}: A feature selection network with realistic synthetic anomaly for anomaly detection.
\newblock In {\em CVPR}, 2024.

\bibitem{zhou2024anomalyclip}
Qihang Zhou, Guansong Pang, Yu~Tian, Shibo He, and Jiming Chen.
\newblock Anomaly{CLIP}: Object-agnostic prompt learning for zero-shot anomaly detection.
\newblock In {\em ICLR}, 2024.

\bibitem{zhou2025pointad}
Qihang Zhou, Jiangtao Yan, Shibo He, Wenchao Meng, and Jiming Chen.
\newblock {PointAD}: Comprehending 3d anomalies from points and pixels for zero-shot 3d anomaly detection.
\newblock In {\em NeurIPS}, 2024.

\bibitem{zhu2024anomaly}
Jiawen Zhu, Choubo Ding, Yu~Tian, and Guansong Pang.
\newblock Anomaly heterogeneity learning for open-set supervised anomaly detection.
\newblock In {\em CVPR}, 2024.

\bibitem{zhu2024toward}
Jiawen Zhu and Guansong Pang.
\newblock Toward generalist anomaly detection via in-context residual learning with few-shot sample prompts.
\newblock In {\em CVPR}, 2024.

\bibitem{VisA}
Yang Zou, Jongheon Jeong, Latha Pemula, Dongqing Zhang, and Onkar Dabeer.
\newblock Spot-the-difference self-supervised pre-training for anomaly detection and segmentation.
\newblock In {\em ECCV}, 2022.

\end{thebibliography}
}

\newpage
\section*{NeurIPS Paper Checklist}

\begin{enumerate}
\item {\bf Claims}
    \item[] Question: Do the main claims made in the abstract and introduction accurately reflect the paper's contributions and scope?
    \item[] Answer: \answerYes{} 
    \item[] Justification: Our contributions are summarized in the \hyperref[sec:abstract]{\textbf{Abstract}} and \hyperref[sec:intro]{\textbf{Introduction}} sections.

\item {\bf Limitations}
    \item[] Question: Does the paper discuss the limitations of the work performed by the authors?
    \item[] Answer: \answerYes{} 
    \item[] Justification: Please see Appendix \ref{sec:limtations}.

\item {\bf Theory assumptions and proofs}
    \item[] Question: For each theoretical result, does the paper provide the full set of assumptions and a complete (and correct) proof?
    \item[] Answer: \answerNA{} 
    \item[] Justification: This paper does not include theoretical results.

    \item {\bf Experimental result reproducibility}
    \item[] Question: Does the paper fully disclose all the information needed to reproduce the main experimental results of the paper to the extent that it affects the main claims and/or conclusions of the paper (regardless of whether the code and data are provided or not)?
    \item[] Answer: \answerYes{} 
    \item[] Justification: Please see the \textbf{Implementation Details} in Sec.~\ref{sec:set_up}.

\item {\bf Open access to data and code}
    \item[] Question: Does the paper provide open access to the data and code, with sufficient instructions to faithfully reproduce the main experimental results, as described in supplemental material?
    \item[] Answer: \answerYes{} 
    \item[] Justification: As mentioned in the Abstract, our code and data will be made publicly available.

\item {\bf Experimental setting/details}
    \item[] Question: Does the paper specify all the training and test details (e.g., data splits, hyperparameters, how they were chosen, type of optimizer, etc.) necessary to understand the results?
    \item[] Answer: \answerYes{} 
    \item[] Justification: Please see Sec.~\ref{sec:set_up} and Appendix \ref{apx:hyper}.

\item {\bf Experiment statistical significance}
    \item[] Question: Does the paper report error bars suitably and correctly defined or other appropriate information about the statistical significance of the experiments?
    \item[] Answer: \answerYes{} 
    \item[] Justification: As mentioned in Sec.~\ref{sec:set_up}, we report results averaged across 3 independent runs with different random seeds. The detailed results can be found in Appendix \ref {apx:details}.

\item {\bf Experiments compute resources}
    \item[] Question: For each experiment, does the paper provide sufficient information on the computer resources (type of compute workers, memory, time of execution) needed to reproduce the experiments?
    \item[] Answer: \answerYes{} 
    \item[] Justification: Please see the \textbf{Implementation Details} in Sec.~\ref{sec:set_up} and \textbf{Efficiency Analysis} at Sec.~\ref{sec:eff}.
    
\item {\bf Code of ethics}
    \item[] Question: Does the research conducted in the paper conform, in every respect, with the NeurIPS Code of Ethics \url{https://neurips.cc/public/EthicsGuidelines}?
    \item[] Answer: \answerYes{} 
    \item[] Justification: We have read the NeurIPS Code of Ethics and confirm that our research complies with these ethical standards.

\item {\bf Broader impacts}
    \item[] Question: Does the paper discuss both potential positive societal impacts and negative societal impacts of the work performed?
    \item[] Answer: \answerYes{} 
    \item[] Justification: Please see the Appendix \ref{sec:limtations}.
    
\item {\bf Safeguards}
    \item[] Question: Does the paper describe safeguards that have been put in place for responsible release of data or models that have a high risk for misuse (e.g., pretrained language models, image generators, or scraped datasets)?
    \item[] Answer: \answerNA{} 
    \item[] Justification: We think that our paper poses no such risks.

\item {\bf Licenses for existing assets}
    \item[] Question: Are the creators or original owners of assets (e.g., code, data, models), used in the paper, properly credited and are the license and terms of use explicitly mentioned and properly respected?
    \item[] Answer: \answerYes{} 
    \item[] Justification: The datasets utilized in this paper are all open-source, and we have cited the corresponding papers accordingly.

\item {\bf New assets}
    \item[] Question: Are new assets introduced in the paper well documented and is the documentation provided alongside the assets?
    \item[] Answer: \answerNA{} 
    \item[] Justification: Our paper introduces a new anomaly detection framework without releasing new assets.

\item {\bf Crowdsourcing and research with human subjects}
    \item[] Question: For crowdsourcing experiments and research with human subjects, does the paper include the full text of instructions given to participants and screenshots, if applicable, as well as details about compensation (if any)? 
    \item[] Answer: \answerNA{} 
    \item[] Justification: Our paper does not involve crowdsourcing nor research with human subjects.

\item {\bf Institutional review board (IRB) approvals or equivalent for research with human subjects}
    \item[] Question: Does the paper describe potential risks incurred by study participants, whether such risks were disclosed to the subjects, and whether Institutional Review Board (IRB) approvals (or an equivalent approval/review based on the requirements of your country or institution) were obtained?
    \item[] Answer: \answerNA{} 
    \item[] Justification: Our paper does not involve crowdsourcing nor research with human subjects.

\item {\bf Declaration of LLM usage}
    \item[] Question: Does the paper describe the usage of LLMs if it is an important, original, or non-standard component of the core methods in this research? Note that if the LLM is used only for writing, editing, or formatting purposes and does not impact the core methodology, scientific rigorousness, or originality of the research, declaration is not required.
    \item[] Answer: \answerNA{} 
    \item[] Justification: The LLM is used only for editing, and the core method development in our paper does not involve LLMs as any important.

\end{enumerate}

\newpage
\appendix

\section{Dataset Collection}
\label{sec:supp_a}
To implement our normal-abnormal-guided generalist anomaly detection, we construct new benchmarks using three popular public datasets, including MVTecAD \cite{MVTec}, VisA \cite{VisA}, and BraTS \cite{menze2014brats}. We conduct evaluations from two perspectives: (1) industrial-to-industrial evaluation across different industrial scenarios (MVTecAD $\leftrightarrow$ VisA), and (2) industrial-to-medical evaluation across industrial and medical domains (MVTecAD $\rightarrow$ BraTS). Specifically, when MVTecAD serves as the training set, we test on VisA and BraTS; when testing on MVTecAD, we use VisA as the training set.

\begin{wraptable}{r}{0.52\textwidth} 
    \vspace{-1em}
    \caption{Statistics of the datasets. $|\mathcal{C}|$ denotes the number of classes, $|\mathcal{T}|$ denotes the number of defect types, $N$ and $A$ represent the normal and abnormal samples, respectively. $K_1$/$K_2$ denotes the number of normal/abnormal reference samples.}
    \label{tab:dataset}%
    \vspace{1em}
    \scriptsize
    \setlength\tabcolsep{3.8pt}
    \renewcommand\arraystretch{0.8}
    \centering
    \begin{tabular}{c|cccccc}
        \toprule
        \multirow{2}[2]{*}{Dataset} & \multirow{2}[2]{*}{$|\mathcal{C}|$} & \multirow{2}[2]{*}{$|\mathcal{T}|$} & \multicolumn{2}{c}{Test} & \multicolumn{2}{c}{Reference} \\
        &       &       & $N$ & $A$ & $N$ & $A$ \\
        \midrule
        MVTecAD & 15    & 73    & 467   & 1258  & $15\times K_1$ & $73\times K_2$ \\
        VisA  & 12    & -     & 962   & 1200  & $12\times K_1$ & $12\times K_2$ \\
        BraTS & 1     & 1     & 154   & 1097  & $1 \times K_1$   & $1\times K_2$ \\
        \bottomrule
    \end{tabular}%
\end{wraptable}%

\vspace{-0.5em}
\paragraph{Original Dataset Structure.} Each dataset is individually divided into its own training and test sets. For all datasets, their respective training sets contain only \textit{good} samples, while each dataset has a unique structure for its corresponding test set. As shown in Tab.~\ref{tab:dataset}. MVTecAD contains $15$ object classes with $73$ different defect types, distinguishing between various defect categories. VisA includes $12$ object classes with various defect types, but all anomalies within each object class are grouped together without specific categorization. BraTS exclusively comprises brain MRI scans, with abnormal samples including \textit{tumour} segmentation. 
\vspace{-0.5em}
\paragraph{Our Input Data.} During both training and testing, we implement a random sampling strategy to construct our input data. Specifically, our input data consists of a query input and a reference set that includes both normal and abnormal samples. We first randomly select a sample from the entire test set as the query input. Then, based on the query input's category, we randomly select $K_1$ normal samples from the training set as normal references. Additionally, according to the query input's \textit{defect-type}, we randomly select $K_2$ abnormal samples as abnormal references. During training, each episode input are randomly combined samples. Unlike the training process, during testing, we randomly selected reference set only once for each anomaly type and then use this set to test all samples of this anomaly type. The detailed statistics of our reference set are shown in Tab.~\ref{tab:dataset}.

\begin{wrapfigure}{r}{0.52\textwidth} 
  \centering
  \includegraphics[width=0.52\textwidth]{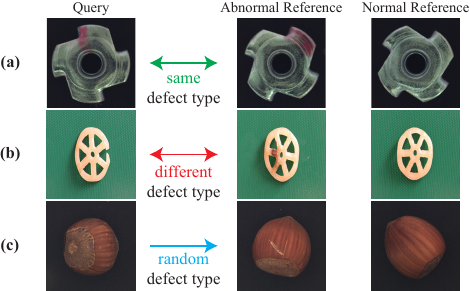}
  \vspace{-0.5em}
  \caption{Some examples of our input data}
  \label{fig:a_dataset}
    \vspace{-0.5em}
\end{wrapfigure}

\paragraph{Defect type between Query and Reference.} As shown in Fig.~\ref{fig:a_dataset}, there are some input cases: (a) the input query and abnormal reference samples belong to the same defect type; (b) the input query and abnormal reference samples exhibit different defect types; (c) when the input query is normal, the abnormal reference randomly selects one defect type. The MVTecAD dataset categorizes different defect types separately, thereby ensuring that the input query ($\mathbf{x}^q$) and abnormal reference samples ($\mathcal{R}^a$) belong to the same defect category (Fig.~\ref{fig:a_dataset}a). In contrast, the VisA dataset combines all anomaly samples for each product type without distinguishing defect types, which results in cases where the input query and abnormal reference may exhibit different types of anomalies (Fig.~\ref{fig:a_dataset}b). Additionally, when the input query is normal, the abnormal reference randomly selects one defect type (Fig.~\ref{fig:a_dataset}c).

\vspace{-0.5em}
\section{Residual Features}
\label{sec:supp_b}
\begin{figure}[hp]
\begin{minipage}[t]{0.49\textwidth}
    \centering
    \includegraphics[width=\linewidth]{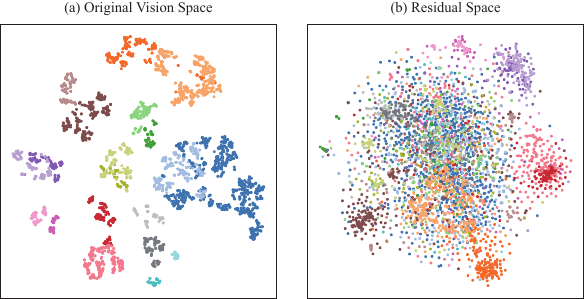}
    \vspace{-1em}
    \caption{T-SNE visualization of features. Different colours denote different defect types. (a) In the original vision space, the features from different defect types are significantly different. (b) In the residual feature space, these features are more overlapped in their distribution.}
    \label{fig:feat_tsne}
    \vspace{-1em}
\end{minipage}
\hfill
\begin{minipage}[t]{0.49\textwidth}
    \centering
    \includegraphics[width=\linewidth]{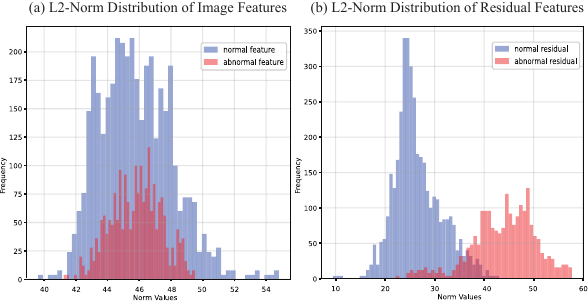}
    \vspace{-1em}
    \caption{(a) In the original visual feature space, the L2-norm distributions of normal and anomalous features exhibit significant overlap. (b) In the residual feature space, the L2-norm distributions of normal and anomalous features are more clearly separated.}
    \label{fig:exp_norm}
    \vspace{-1em}
\end{minipage}
\end{figure}

\vspace{-0.5em}
To explore the guidance of abnormal residual features in anomaly detection, we visualize the feature distribution in the original vision space and the residual feature space. As shown in Fig. \ref{fig:feat_tsne}, the features from different defect types are significantly different in the original vision space, while these features are more concentrated in their distribution in the residual feature space, suggesting it provides a unified cross-domain representation for anomalies. Moreover, as shown in Fig. \ref{fig:exp_norm}, the L2-norm of normal and abnormal features are overlapped in the original vision space, while they are separated in the residual feature space, indicating that residual features offer a more discriminative attribute for distinguishing between normal and abnormal samples.

\vspace{-1em}
\section{More Abnormal References}
\label{sec:supp_c}
\vspace{-1em}
\begin{wraptable}{r}{0.5\textwidth}
\vspace{-1.5em}
  \caption{The AUROC performance comparison between $1$ and $2$ abnormal references  ($A^1$ and $A^2$).}
  \label{tab:2_abnormal}%
  \vspace{1em}
  \scriptsize
    \setlength\tabcolsep{8pt}
  \renewcommand\arraystretch{0.8}
  \centering
    \begin{tabular}{c|cccc}
    \toprule
    Setting & MVTecAD & VisA  & Mean  & $\Delta$ \\
    \midrule
    $N^1+A^1$ & 95.8  & 88.5  & 92.2  &  \\
    $N^1+A^2$ & 96.5  & 89.5  & 93.0  & \textcolor[rgb]{ 1,  0,  0}{+ 0.8} \\
    \midrule
    \midrule
    $N^2+A^1$ & 96.8  & 89.8  & 93.3  &  \\
    $N^2+A^2$ & 97.3  & 90.5  & 93.9  & \textcolor[rgb]{ 1,  0,  0}{+ 0.6} \\
    \midrule
    \midrule
    $N^4+A^1$ & 97.1  & 91.2  & 94.2  &  \\
    $N^4+A^2$ & 97.5  & 92.2  & 94.9  & \textcolor[rgb]{ 1,  0,  0}{+ 0.7} \\
    \bottomrule
    \end{tabular}%
\end{wraptable}%
Considering the scarcity of anomalous samples in real-world scenarios, we focused on using only one anomalous reference sample in the main text. To further explore the scalability of the NAGL framework, we conducted experiments using two anomalous reference samples. As shown in Table \ref{tab:2_abnormal}, using two anomalous reference samples improved the performance of the NAGL framework on both MVTecAD and VisA datasets. This shows NAGL effectively uses multiple anomalous references to improve detection performance.

\vspace{-0.5em}
\section{Compared with ViT-based Results}
\label{sec:supp_d}
\vspace{-0.5em}
Due to the extensive training time required for ResAD, we only reproduced the CNN-based (WideResNet50) results in the main text, maintaining identical training hyperparameters as specified in the original paper. As shown in Tab.~\ref{tab:suppl_vit}, for a more equitable comparison, we also compare our approach with ViT-based results. It is clear that our method still achieves competitive performance.

\begin{table}[hp]
\vspace{-0.5em}
  \caption{The AUROC performance comparison of ViT-based backbones on MVTecAD and VisA datasets.}
  \label{tab:suppl_vit}%
  \scriptsize
\setlength\tabcolsep{5pt}
\renewcommand\arraystretch{0.8}
  \centering
    \begin{tabular}{cc|ccccrcc|cccc}
        \cmidrule{1-6}\cmidrule{8-13}    \multirow{2}[2]{*}{Setting} & \multirow{2}[2]{*}{Method} & \multicolumn{2}{c}{MVTecAD} & \multicolumn{2}{c}{VisA} &       & \multirow{2}[2]{*}{Setting} & \multirow{2}[2]{*}{Method} & \multicolumn{2}{c}{MVTecAD} & \multicolumn{2}{c}{VisA} \\
                  &       & Image & Pixel & Image & Pixel &       &       &       & Image & Pixel & Image & Pixel \\
        \cmidrule{1-6}\cmidrule{8-13}    \multirow{3}[2]{*}{$N^2$} & WinCLIP & 94.4  & 96.0  & 84.6  & 96.8  &       & \multirow{3}[2]{*}{$N^4$} & WinCLIP & 95.2  & 96.2  & 87.3  & 97.2  \\
                  & InCTRL & 94.0  & -     & 85.8  & -     &       &       & InCTRL & 94.5  & -     & 87.7  & - \\
                  & ResAD & 94.4  & 95.6  & 84.5  & 95.1  &       &       & ResAD & 94.2  & 96.9  & 90.8  & 97.5  \\
        \cmidrule{1-6}\cmidrule{8-13}    $N^2+A^1$ & Ours  & 96.8  & 96.8  & 89.8  & 97.6  &       & $N^4+A^1$ & Ours  & 97.1  & 97.0  & 91.2  & 97.8  \\
        \cmidrule{1-6}\cmidrule{8-13}    
    \end{tabular}%
    \vspace{-0.5em}
\end{table}%
\vspace{-0.5em}
\section{Discussion}
\label{sec:supp_e}
\vspace{-0.5em}
\label{sec:limtations}
\textbf{Limitations:} A drawback of our study is that it focuses solely on image data for experimentation. It would be highly beneficial to apply our approach to other data types, like video and time series, to thoroughly assess the adaptability of our method. 

\textbf{Social Impacts:} As a unified framework for generalist anomaly detection, the introduced approach does not raise specific ethical issues or adverse societal effects. The datasets utilized are publicly available. All qualitative illustrations are derived from industrial product imagery, ensuring no violation of personal privacy.

\section{Hyperparameter Analysis}
\label{sec:supp_f}
\label{apx:hyper}
\begin{wrapfigure}{r}{0.5\textwidth}
    \centering
    \includegraphics[width=1.0\linewidth]{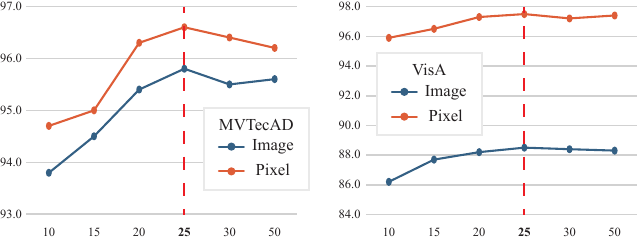}
    \vspace{-1em}
    \caption{Impact of the number of learnable proxies $M$ on MVTecAD and VisA datasets.}
    \label{fig:exp_hyper}
\end{wrapfigure}
Fig.~\ref{fig:exp_hyper} shows the impact of the number of learnable proxies $M$. We observe that the performance of our method increases as $M$ increases. This is because a larger number of proxies can better capture the diversity of the feature space, which is beneficial for the refinement process. However, the performance improvement diminishes as $M$ exceeds $25$. One possible reason is the model becomes over-parameterized, leading to overfitting. Therefore, we set $M=25$ in our experiments to balance performance and efficiency.

\section{Additional Results}
\label{sec:supp_g}
\label{apx:details}
Additionally, we evaluated the performance of our proposed method on the MVTec3D and MVTecLOCO datasets to further investigate its generalizability. As shown in Tab.~\ref{tab:exp_loco}, our method achieves competitive results compared to ResAD in both 3D anomaly detection and logical anomaly detection, demonstrating its generalizability.

\begin{table}[hp]
    \caption{Comparison of the proposed method with the ResAD on MVTec3D and MVTecLOCO datasets.} 
    \label{tab:exp_loco}%
    \scriptsize
    \setlength\tabcolsep{3.0pt}
    \centering
    \begin{tabular}{cc|cccccc|cccccc}
    \toprule
    \multirow{3}[2]{*}{Setting} & \multirow{3}[2]{*}{Method} & \multicolumn{6}{c|}{\textbf{MVTec3D}}         & \multicolumn{6}{c}{\textbf{MVTecLOCO}} \\
          &       & \multicolumn{3}{c}{Image-level} & \multicolumn{3}{c|}{Pixel-level} & \multicolumn{3}{c}{Image-level} & \multicolumn{3}{c}{Pixel-level} \\
          &       & AUROC & AP    & F1-max & AUROC & PRO   & F1-max & AUROC & AP    & F1-max & AUROC & PRO   & F1-max \\
    \midrule
    $N^1$ & ResAD$^{\ast}$ & 63.8  & 86.4  & 88.7  & 94.2  & 88.2  & 21.2  & 60.6  & 73.9  & 77.8  & 65.9  & 60.8  & 17.9  \\
    $N^1+A^1$ & Ours  & \textbf{83.0} & \textbf{95.0} & \textbf{91.7} & \textbf{94.7} & \textbf{98.4} & \textbf{50.3} & \textbf{63.3} & \textbf{76.3} & \textbf{78.4} & \textbf{66.6} & \textbf{69.9} & \textbf{19.4} \\
    \midrule
    \midrule
    $N^2$ & ResAD$^{\ast}$ & 66.7  & 88.6  & 88.9  & \textbf{94.9} & 90.1  & 24.5  & 62.3  & 76.3  & 77.7  & 66.0  & 62.3  & 18.3  \\
    $N^2+A^1$ & Ours  & \textbf{82.8} & \textbf{94.9} & \textbf{91.6} & 94.8  & \textbf{98.4} & \textbf{51.3} & \textbf{64.6} & \textbf{77.4} & \textbf{78.0} & \textbf{66.3} & \textbf{69.3} & \textbf{18.5} \\
    \midrule
    \midrule
    $N^4$ & ResAD$^{\ast}$ & 70.1  & 89.4  & 88.9  & 95.0  & 91.4  & 26.0  & 65.7  & 77.5  & 77.7  & 67.5  & 60.7  & \textbf{19.1} \\
    $N^4+A^1$ & Ours  & \textbf{86.9} & \textbf{96.4} & \textbf{92.3} & \textbf{95.4} & \textbf{98.6} & \textbf{54.2} & \textbf{71.5} & \textbf{81.5} & \textbf{79.1} & \textbf{66.9} & \textbf{69.2} & 18.6  \\
    \bottomrule
    \end{tabular}%
  \end{table}%

\begin{figure}[hp]
    \centering
    \includegraphics[width=\linewidth]{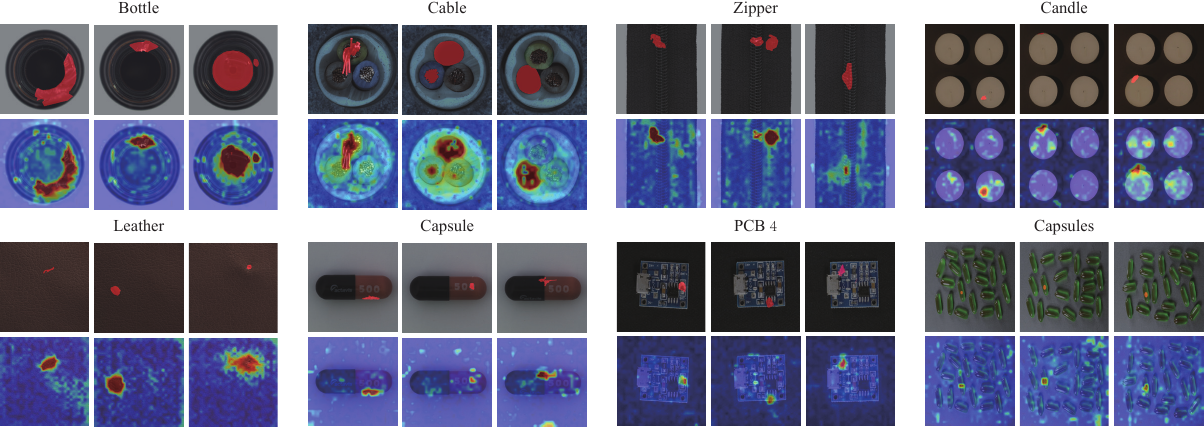}
    \caption{Qualitative results for eight subcategories with different defect types.}
    \label{fig:suppl_vis}
\end{figure}

As shown in Fig.~\ref{fig:suppl_vis}, we provide further qualitative results obtained from our NAGL for pixel-level anomaly detection. The results demonstrate that our approach accurately localizes both large and small surface defects across various test cases. Furthermore, we report the detailed subset-level results ($mean\pm std$) of NAGL on MVTecAD, VisA, BraTS, MVTec3D and MVTecLOCO datasets, under $N^1+A^1$ (Tab.~\ref{tab:full_results_1}), $N^2+A^1$ (Tab.~\ref{tab:full_results_2}), and $N^4+A^1$ (Tab.~\ref{tab:full_results_3}) settings. We also report performance on the BTAD and MPDD datasets.

\begin{table}[hp]
    \caption{The comprehensive results of our NAGL framework, evaluated under settings with $N^1+A^1$.}
    \label{tab:full_results_1}%
    \scriptsize
    \setlength\tabcolsep{8pt}
    \centering
    \begin{tabular}{cc|ccc|ccc}
    \toprule
    \multirow{2}[2]{*}{Dataset} & \multirow{2}[2]{*}{Objects} & \multicolumn{3}{c|}{Image-level} & \multicolumn{3}{c}{Pixel-level} \\
          &       & AUROC & AP    & F1-max & AUROC & PRO   & F1-max \\
    \midrule
    \multicolumn{1}{c}{\multirow{16}[4]{*}{\textbf{MVTecAD}}} & bottle & 99.2±0.9 & 99.7±0.4 & 99.2±0.0 & 98.2±0.1 & 95.8±0.2 & 76.2±0.8 \\
          & cable & 93.1±0.9 & 96.3±0.6 & 90.4±2.0 & 94.2±0.6 & 87.8±0.5 & 58.9±0.8 \\
          & capsule & 91.5±7.1 & 98.1±1.7 & 95.5±0.7 & 98.1±0.2 & 96.7±0.4 & 46.8±5.3 \\
          & carpet & 100.0±0.0 & 100.0±0.0 & 100.0±0.0 & 99.3±0.0 & 98.5±0.0 & 68.9±0.6 \\
          & grid  & 99.8±0.2 & 99.9±0.1 & 99.4±0.5 & 99.4±0.0 & 96.9±0.3 & 51.4±0.9 \\
          & hazelnut & 99.5±0.4 & 99.7±0.2 & 98.3±1.1 & 99.5±0.1 & 97.1±0.4 & 77.5±2.7 \\
          & leather & 100.0±0.0 & 100.0±0.0 & 100.0±0.0 & 99.1±0.1 & 98.5±0.3 & 41.3±1.4 \\
          & metal\_nut & 99.4±0.1 & 99.9±0.0 & 98.2±0.3 & 95.2±0.3 & 93.0±0.5 & 68.5±1.2 \\
          & pill  & 93.6±1.6 & 98.7±0.3 & 95.9±0.5 & 94.2±0.3 & 96.2±0.1 & 52.9±2.3 \\
          & screw & 77.2±0.9 & 91.5±0.4 & 86.7±0.2 & 97.8±0.4 & 92.2±0.8 & 43.2±2.4 \\
          & tile  & 99.8±0.1 & 99.9±0.0 & 99.0±0.3 & 96.1±0.4 & 91.4±0.3 & 70.1±0.4 \\
          & toothbrush & 99.4±0.2 & 99.8±0.0 & 98.3±0.0 & 99.2±0.1 & 95.1±0.2 & 65.9±2.2 \\
          & transistor & 86.1±4.2 & 79.3±6.9 & 76.7±4.1 & 84.3±3.7 & 64.7±1.7 & 39.1±6.4 \\
          & wood  & 99.8±0.1 & 99.9±0.0 & 98.9±0.5 & 96.0±0.9 & 95.8±0.5 & 67.9±0.7 \\
          & zipper & 99.1±0.2 & 99.8±0.1 & 98.5±0.5 & 97.9±0.1 & 94.0±0.3 & 54.5±0.6 \\
\cmidrule{2-8}          & \textbf{Mean} & \textbf{95.8±0.7} & \textbf{97.5±0.6} & \textbf{95.7±0.3} & \textbf{96.6±0.2} & \textbf{92.9±0.0} & \textbf{58.9±0.5} \\
    \midrule
    \midrule
    \multicolumn{1}{c}{\multirow{13}[4]{*}{\textbf{VisA}}} & candle & 93.9±0.7 & 94.3±0.3 & 87.3±1.0 & 99.0±0.0 & 97.1±0.3 & 37.6±0.9 \\
          & capsules & 96.7±0.4 & 98.0±0.2 & 93.5±0.1 & 97.9±0.4 & 96.4±0.4 & 44.4±4.4 \\
          & cashew & 88.5±2.3 & 94.3±0.9 & 87.2±1.8 & 98.7±0.2 & 97.6±1.1 & 56.7±1.5 \\
          & chewinggum & 97.3±0.3 & 98.9±0.1 & 95.7±0.3 & 99.3±0.1 & 84.6±0.6 & 68.7±0.4 \\
          & fryum & 95.7±0.7 & 98.1±0.3 & 92.8±0.9 & 95.8±0.4 & 91.1±0.4 & 40.8±2.4 \\
          & macaroni1 & 89.2±2.6 & 88.7±3.1 & 83.1±2.3 & 99.4±0.0 & 94.1±1.0 & 27.6±2.9 \\
          & macaroni2 & 59.5±7.5 & 55.9±6.3 & 68.3±1.0 & 98.2±0.6 & 86.2±2.5 & 12.6±4.2 \\
          & pcb1  & 94.9±0.7 & 93.4±0.7 & 90.6±1.2 & 98.9±0.2 & 94.2±0.1 & 60.7±4.6 \\
          & pcb2  & 86.3±0.9 & 84.3±1.4 & 80.0±1.4 & 95.6±0.1 & 86.1±0.3 & 35.5±1.0 \\
          & pcb3  & 85.6±2.1 & 88.5±1.6 & 78.5±0.7 & 93.7±0.1 & 86.6±0.1 & 42.0±2.5 \\
          & pcb4  & 79.1±11.1 & 80.3±7.4 & 76.3±9.7 & 94.9±0.6 & 81.9±3.7 & 25.7±2.2 \\
          & pipe\_fryum & 95.8±0.6 & 98.0±0.3 & 91.9±1.0 & 98.8±0.1 & 97.8±0.0 & 48.9±2.2 \\
\cmidrule{2-8}          & \textbf{Mean} & \textbf{88.5±1.4} & \textbf{89.4±1.2} & \textbf{85.4±0.7} & \textbf{97.5±0.1} & \textbf{91.1±0.4} & \textbf{41.8±0.5} \\
    \midrule
    \midrule
    \textbf{BraTS} & \textbf{brain} & \textbf{78.1±3.4} & \textbf{95.4±0.9} & \textbf{93.8±0.3} & \textbf{96.5±0.1} & \textbf{79.4±0.5} & \textbf{49.4±1.0} \\
    \midrule
    \midrule
    \multicolumn{1}{c}{\multirow{11}[4]{*}{\textbf{MVTec3D}}} & bagel & 95.5±2.0 & 98.9±0.5 & 95.9±1.6 & 98.9±0.1 & 99.7±0.1 & 66.9±0.8 \\
          & cable\_gland & 77.2±11.1 & 93.4±4.2 & 90.4±0.7 & 95.6±2.2 & 98.4±0.8 & 38.5±20.1 \\
          & carrot & 87.4±5.0 & 96.8±1.5 & 93.0±1.3 & 99.1±0.2 & 99.8±0.0 & 54.2±2.8 \\
          & cookie & 83.9±6.1 & 95.3±1.8 & 90.5±1.9 & 92.9±0.3 & 97.7±0.2 & 57.7±4.2 \\
          & dowel & 74.8±2.7 & 92.9±1.2 & 90.7±1.0 & 94.3±0.5 & 98.6±0.2 & 40.0±3.5 \\
          & foam  & 78.0±2.2 & 94.6±0.5 & 88.9±0.0 & 79.8±1.2 & 92.8±0.7 & 43.5±1.7 \\
          & peach & 93.8±1.8 & 98.0±0.9 & 94.9±0.8 & 99.2±0.1 & 99.8±0.0 & 62.5±4.4 \\
          & potato & 67.7±8.5 & 89.7±3.6 & 89.9±1.0 & 98.1±0.3 & 99.5±0.1 & 42.5±4.6 \\
          & rope  & 96.2±1.3 & 98.6±0.4 & 94.7±0.8 & 97.7±0.0 & 99.4±0.0 & 50.4±0.2 \\
          & tire  & 75.3±5.8 & 92.5±1.8 & 87.9±0.4 & 91.9±0.3 & 98.2±0.1 & 46.4±0.9 \\
\cmidrule{2-8}          & \textbf{Mean} & \textbf{83.0±1.5} & \textbf{95.0±0.6} & \textbf{91.7±0.4} & \textbf{94.7±0.2} & \textbf{98.4±0.1} & \textbf{50.3±2.4} \\
    \midrule
    \midrule
    \multicolumn{1}{c}{\multirow{6}[2]{*}{\textbf{MVTecLOCO}}} & breakfast\_box & 77.1±2.8 & 87.1±1.7 & 79.3±1.8 & 67.1±1.6 & 79.7±1.2 & 32.3±0.9 \\
          & juice\_bottle & 50.8±2.1 & 74.6±1.9 & 83.5±0.3 & 71.4±0.5 & 85.9±0.4 & 30.8±1.7 \\
          & pushpins & 62.7±0.6 & 66.4±1.4 & 73.5±1.9 & 53.1±1.4 & 54.3±1.4 & 3.1±0.1 \\
          & screw\_bag & 59.7±1.1 & 75.8±1.0 & 78.5±0.3 & 67.8±1.6 & 63.7±1.2 & 10.8±0.1 \\
          & splicing\_connectors & 66.2±5.3 & 77.5±6.1 & 76.9±0.4 & 73.7±1.3 & 65.7±3.5 & 20.0±1.9 \\
          & \textbf{Mean} & \textbf{63.3±1.2} & \textbf{76.3±1.0} & \textbf{78.4±0.5} & \textbf{66.6±0.3} & \textbf{69.9±1.0} & \textbf{19.4±0.2} \\
    \midrule
    \midrule
    \multicolumn{1}{c}{\multirow{4}[4]{*}{\textbf{BTAD}}} & 01    & 85.9±3.0 & 95.0±1.1 & 86.9±2.0 & 65.4±1.7 & 93.3±1.0 & 50.9±2.2 \\
          & 02    & 93.6±1.4 & 99.0±0.2 & 95.6±0.6 & 68.4±5.8 & 96.9±0.4 & 64.6±3.2 \\
          & 03    & 99.8±0.0 & 96.7±0.5 & 93.2±2.1 & 97.3±0.3 & 99.5±0.0 & 65.6±1.6 \\
\cmidrule{2-8}          & \textbf{Mean} & \textbf{93.1±1.3} & \textbf{96.9±0.5} & \textbf{91.9±1.3} & \textbf{77.1±2.0} & \textbf{96.6±0.3} & \textbf{60.4±0.7} \\
    \midrule
    \midrule
    \multicolumn{1}{c}{\multirow{7}[4]{*}{\textbf{MPDD}}} & bracket\_black & 55.7±2.9 & 66.6±6.5 & 76.0±1.6 & 92.0±1.2 & 95.3±0.6 & 23.1±5.8 \\
          & bracket\_brown & 57.6±4.2 & 75.0±3.8 & 81.6±0.7 & 87.0±1.3 & 94.9±0.2 & 10.9±0.3 \\
          & bracket\_white & 76.2±2.8 & 68.9±7.3 & 84.6±2.1 & 95.9±0.7 & 99.7±0.0 & 24.7±4.1 \\
          & connector & 87.1±3.2 & 70.9±6.1 & 76.4±3.9 & 91.2±1.5 & 97.4±0.5 & 28.8±4.7 \\
          & metal\_plate & 99.9±0.0 & 100.0±0.0 & 99.3±0.0 & 89.7±0.9 & 96.5±0.5 & 73.2±1.7 \\
          & tubes & 94.5±1.1 & 97.8±0.6 & 92.0±0.5 & 95.5±0.5 & 98.8±0.1 & 64.6±1.5 \\
\cmidrule{2-8}          & \textbf{Mean} & \textbf{78.5±0.8} & \textbf{79.9±1.1} & \textbf{85.0±0.5} & \textbf{91.9±0.6} & \textbf{97.1±0.2} & \textbf{37.5±2.5} \\
    \bottomrule
    \end{tabular}%
\end{table}%

\begin{table}[hp]
    \caption{The comprehensive results of our NAGL framework, evaluated under settings with $N^2+A^1$.}
    \label{tab:full_results_2}%
    \scriptsize
    \setlength\tabcolsep{8pt}
    \centering
    \begin{tabular}{cc|ccc|ccc}
    \toprule
    \multirow{2}[2]{*}{Dataset} & \multirow{2}[2]{*}{Objects} & \multicolumn{3}{c|}{Image-level} & \multicolumn{3}{c}{Pixel-level} \\
          &       & AUROC & AP    & F1-max & AUROC & PRO   & F1-max \\
    \midrule
    \multicolumn{1}{c}{\multirow{16}[4]{*}{\textbf{MVTecAD}}} & bottle & 99.2±1.0 & 99.7±0.4 & 99.5±0.5 & 98.2±0.2 & 96.0±0.5 & 76.5±1.4 \\
          & cable & 93.7±0.4 & 96.6±0.3 & 90.8±0.5 & 94.7±0.1 & 88.5±0.1 & 60.0±1.4 \\
          & capsule & 90.8±8.3 & 97.8±2.1 & 95.4±1.3 & 98.1±0.1 & 96.8±0.5 & 47.4±5.2 \\
          & carpet & 100.0±0.0 & 100.0±0.0 & 100.0±0.0 & 99.3±0.0 & 98.5±0.0 & 68.9±0.4 \\
          & grid  & 99.9±0.1 & 100.0±0.0 & 99.7±0.5 & 99.4±0.0 & 97.0±0.0 & 51.3±0.8 \\
          & hazelnut & 99.8±0.1 & 99.9±0.1 & 99.3±0.0 & 99.6±0.1 & 97.4±0.4 & 79.8±1.0 \\
          & leather & 100.0±0.0 & 100.0±0.0 & 100.0±0.0 & 99.1±0.0 & 98.4±0.3 & 41.8±1.0 \\
          & metal\_nut & 99.8±0.1 & 99.9±0.0 & 99.3±0.3 & 95.5±0.3 & 93.5±0.5 & 70.2±1.1 \\
          & pill  & 95.5±1.6 & 99.1±0.4 & 96.6±0.2 & 94.3±0.3 & 96.2±0.2 & 54.4±1.1 \\
          & screw & 86.0±0.9 & 95.2±0.4 & 88.6±1.6 & 98.2±0.2 & 93.5±0.3 & 48.8±0.4 \\
          & tile  & 99.9±0.1 & 100.0±0.0 & 99.6±0.3 & 96.1±0.3 & 91.4±0.4 & 69.9±0.4 \\
          & toothbrush & 99.4±0.7 & 99.7±0.3 & 98.4±1.6 & 99.2±0.1 & 95.5±0.2 & 65.5±2.0 \\
          & transistor & 88.0±4.1 & 80.8±4.6 & 79.2±4.7 & 85.2±1.2 & 66.0±1.8 & 39.0±4.0 \\
          & wood  & 99.8±0.1 & 99.9±0.0 & 99.2±0.0 & 96.4±0.8 & 95.9±0.3 & 68.8±1.4 \\
          & zipper & 99.5±0.2 & 99.9±0.0 & 98.9±0.2 & 98.0±0.2 & 94.1±0.4 & 54.7±0.7 \\
\cmidrule{2-8}          & \textbf{Mean} & \textbf{96.8±0.8} & \textbf{97.9±0.4} & \textbf{96.3±0.5} & \textbf{96.8±0.2} & \textbf{93.2±0.2} & \textbf{59.8±0.3} \\
    \midrule
    \midrule
    \multicolumn{1}{c}{\multirow{13}[4]{*}{\textbf{VisA}}} & candle & 93.7±0.7 & 94.2±0.5 & 87.9±0.5 & 99.0±0.0 & 97.1±0.2 & 37.9±1.2 \\
          & capsules & 96.9±1.0 & 98.2±0.5 & 93.4±1.7 & 97.9±0.2 & 96.5±0.4 & 48.0±3.7 \\
          & cashew & 90.1±3.6 & 95.3±1.7 & 88.9±2.6 & 98.7±0.1 & 97.8±0.8 & 57.2±0.5 \\
          & chewinggum & 97.7±0.5 & 99.0±0.2 & 95.5±0.4 & 99.3±0.1 & 85.1±1.3 & 68.1±0.7 \\
          & fryum & 95.9±0.1 & 98.2±0.1 & 94.0±0.2 & 95.7±0.2 & 91.1±0.3 & 41.3±1.2 \\
          & macaroni1 & 89.1±0.3 & 88.5±0.8 & 83.4±0.5 & 99.4±0.0 & 93.8±0.6 & 27.1±1.8 \\
          & macaroni2 & 63.2±5.4 & 59.3±4.3 & 68.8±1.4 & 98.3±0.2 & 86.7±0.4 & 17.3±3.8 \\
          & pcb1  & 94.8±0.8 & 92.8±1.2 & 91.8±0.4 & 99.0±0.1 & 94.3±0.1 & 61.1±1.8 \\
          & pcb2  & 86.6±1.6 & 84.4±1.4 & 81.5±2.4 & 95.7±0.1 & 86.5±0.1 & 36.9±0.8 \\
          & pcb3  & 88.9±1.1 & 91.2±0.7 & 83.1±1.8 & 93.9±0.1 & 87.3±0.0 & 44.1±2.5 \\
          & pcb4  & 84.6±10.7 & 87.4±8.3 & 80.0±9.6 & 95.2±0.5 & 83.2±4.0 & 29.7±1.8 \\
          & pipe\_fryum & 96.5±1.2 & 98.3±0.6 & 93.1±0.8 & 98.9±0.1 & 97.8±0.1 & 50.7±1.5 \\
\cmidrule{2-8}          & \textbf{Mean} & \textbf{89.8±1.5} & \textbf{90.6±0.9} & \textbf{86.8±1.1} & \textbf{97.6±0.0} & \textbf{91.4±0.4} & \textbf{43.3±0.3} \\
    \midrule
    \midrule
    \textbf{BraTS} & \textbf{brain} & \textbf{82.1±3.0} & \textbf{96.6±1.0} & \textbf{93.9±0.1} & \textbf{96.8±0.2} & \textbf{79.9±0.4} & \textbf{51.8±1.5} \\
    \midrule
    \midrule
    \multicolumn{1}{c}{\multirow{11}[4]{*}{\textbf{MVTec3D}}} & bagel & 95.6±2.6 & 98.9±0.7 & 95.3±1.8 & 99.0±0.1 & 99.6±0.0 & 67.8±1.0 \\
          & cable\_gland & 82.0±4.7 & 95.4±1.1 & 91.0±1.2 & 96.0±1.4 & 98.4±0.5 & 48.5±2.3 \\
          & carrot & 87.1±1.9 & 96.9±0.5 & 92.4±1.2 & 99.1±0.1 & 99.8±0.0 & 55.3±2.9 \\
          & cookie & 81.1±6.5 & 94.4±1.9 & 90.0±1.8 & 92.4±0.7 & 97.6±0.2 & 56.9±3.4 \\
          & dowel & 74.3±2.2 & 92.4±1.3 & 90.6±0.5 & 94.8±0.4 & 98.8±0.1 & 40.7±2.6 \\
          & foam  & 80.5±2.1 & 95.2±0.5 & 89.4±0.5 & 79.9±0.7 & 92.9±0.4 & 43.7±0.6 \\
          & peach & 94.2±2.5 & 98.2±1.1 & 94.8±1.4 & 99.2±0.1 & 99.8±0.0 & 62.8±3.7 \\
          & potato & 67.4±2.8 & 89.7±1.4 & 89.5±0.4 & 98.2±0.1 & 99.6±0.0 & 42.7±3.3 \\
          & rope  & 96.4±1.3 & 98.7±0.4 & 95.3±0.5 & 97.7±0.1 & 99.4±0.0 & 50.4±0.3 \\
          & tire  & 69.3±12.0 & 89.2±6.2 & 87.6±0.3 & 91.7±0.9 & 98.1±0.2 & 44.1±5.3 \\
\cmidrule{2-8}          & \textbf{Mean} & \textbf{82.8±2.9} & \textbf{94.9±1.1} & \textbf{91.6±0.6} & \textbf{94.8±0.3} & \textbf{98.4±0.1} & \textbf{51.3±1.7} \\
    \midrule
    \midrule
    \multicolumn{1}{c}{\multirow{6}[2]{*}{\textbf{MVTecLOCO}}} & breakfast\_box & 78.1±2.1 & 88.3±0.9 & 78.8±1.5 & 64.4±1.9 & 77.7±1.5 & 29.9±1.9 \\
          & juice\_bottle & 58.8±3.7 & 79.2±1.6 & 83.9±0.8 & 70.9±0.7 & 85.6±0.4 & 29.4±1.0 \\
          & pushpins & 57.0±2.1 & 63.3±4.4 & 72.0±0.5 & 51.8±0.8 & 53.9±1.2 & 3.0±0.3 \\
          & screw\_bag & 60.0±2.7 & 76.2±2.0 & 78.2±0.0 & 69.2±0.6 & 65.2±1.3 & 10.9±0.1 \\
          & splicing\_connectors & 69.4±4.8 & 80.2±5.0 & 77.2±0.4 & 75.0±0.7 & 64.3±2.1 & 19.3±0.8 \\
          & \textbf{Mean} & \textbf{64.6±1.4} & \textbf{77.4±1.7} & \textbf{78.0±0.2} & \textbf{66.3±0.4} & \textbf{69.3±0.4} & \textbf{18.5±0.5} \\
    \midrule
    \midrule
    \multicolumn{1}{c}{\multirow{4}[4]{*}{\textbf{BTAD}}} & 01    & 86.4±0.2 & 95.0±0.1 & 85.9±0.6 & 65.6±2.7 & 93.5±1.1 & 52.1±1.3 \\
          & 02    & 93.1±0.7 & 98.9±0.1 & 95.6±0.2 & 65.5±2.0 & 96.9±0.1 & 65.2±1.6 \\
          & 03    & 99.7±0.1 & 95.5±1.8 & 91.8±2.8 & 97.2±0.0 & 99.5±0.0 & 64.6±0.5 \\
\cmidrule{2-8}          & \textbf{Mean} & \textbf{93.0±0.3} & \textbf{96.5±0.5} & \textbf{91.1±0.9} & \textbf{76.1±1.4} & \textbf{96.7±0.4} & \textbf{60.6±1.0} \\
    \midrule
    \midrule
    \multicolumn{1}{c}{\multirow{7}[4]{*}{\textbf{MPDD}}} & bracket\_black & 64.6±8.6 & 73.6±7.7 & 77.8±2.4 & 94.0±0.9 & 96.5±0.6 & 26.5±3.8 \\
          & bracket\_brown & 60.8±1.4 & 75.6±1.6 & 82.0±1.0 & 89.0±0.8 & 95.4±0.5 & 13.1±1.1 \\
          & bracket\_white & 81.1±6.7 & 77.1±10.9 & 84.7±1.0 & 96.5±1.3 & 99.8±0.1 & 23.3±4.5 \\
          & connector & 87.8±4.0 & 72.6±6.3 & 77.8±2.1 & 92.0±1.6 & 97.6±0.5 & 30.7±4.5 \\
          & metal\_plate & 99.9±0.1 & 100.0±0.0 & 99.5±0.4 & 90.0±0.7 & 96.6±0.4 & 73.5±1.6 \\
          & tubes & 94.0±0.6 & 97.6±0.3 & 91.8±0.8 & 95.6±0.5 & 98.8±0.1 & 64.9±0.8 \\
\cmidrule{2-8}          & \textbf{Mean} & \textbf{81.4±2.9} & \textbf{82.7±3.0} & \textbf{85.6±0.3} & \textbf{92.9±0.5} & \textbf{97.5±0.1} & \textbf{38.7±1.9} \\
    \bottomrule
    \end{tabular}%
\end{table}%

\begin{table}[hp]
    \caption{The comprehensive results of our NAGL framework, evaluated under settings with $N^4+A^1$.}
    \label{tab:full_results_3}%
    \setlength\tabcolsep{8pt}
    \scriptsize
    \centering
    \begin{tabular}{cc|ccc|ccc}
    \toprule
    \multirow{2}[2]{*}{Dataset} & \multirow{2}[2]{*}{Objects} & \multicolumn{3}{c|}{Image-level} & \multicolumn{3}{c}{Pixel-level} \\
          &       & AUROC & AP    & F1-max & AUROC & PRO   & F1-max \\
    \midrule
    \multicolumn{1}{c}{\multirow{16}[4]{*}{\textbf{MVTecAD}}} & bottle & 99.8±0.2 & 99.9±0.1 & 99.2±0.0 & 98.1±0.2 & 95.7±0.2 & 76.6±0.7 \\
          & cable & 94.7±0.7 & 97.2±0.3 & 91.2±0.5 & 95.2±0.3 & 89.4±0.8 & 61.0±1.2 \\
          & capsule & 97.4±1.0 & 99.5±0.2 & 97.3±0.8 & 98.4±0.0 & 97.3±0.1 & 52.4±0.4 \\
          & carpet & 100.0±0.0 & 100.0±0.0 & 100.0±0.0 & 99.3±0.0 & 98.4±0.0 & 68.5±0.2 \\
          & grid  & 99.8±0.3 & 99.9±0.1 & 99.7±0.5 & 99.4±0.0 & 97.0±0.1 & 51.3±0.9 \\
          & hazelnut & 99.8±0.2 & 99.9±0.1 & 99.3±0.7 & 99.6±0.1 & 97.5±0.7 & 79.7±1.5 \\
          & leather & 100.0±0.0 & 100.0±0.0 & 100.0±0.0 & 99.1±0.0 & 98.2±0.3 & 41.0±1.1 \\
          & metal\_nut & 99.8±0.0 & 100.0±0.0 & 99.1±0.3 & 96.0±0.3 & 94.0±0.1 & 72.6±1.1 \\
          & pill  & 95.3±1.4 & 99.1±0.3 & 96.3±0.5 & 95.3±0.2 & 96.2±0.2 & 55.0±0.9 \\
          & screw & 84.4±4.2 & 94.5±1.8 & 88.4±0.9 & 98.3±0.3 & 93.8±0.7 & 47.0±1.9 \\
          & tile  & 100.0±0.0 & 100.0±0.0 & 99.6±0.3 & 96.0±0.3 & 91.2±0.2 & 69.5±0.5 \\
          & toothbrush & 99.8±0.2 & 99.9±0.1 & 98.9±0.9 & 99.2±0.0 & 95.5±0.5 & 64.7±1.4 \\
          & transistor & 85.7±5.9 & 79.9±7.0 & 79.0±4.4 & 85.8±1.2 & 67.8±1.7 & 38.4±0.9 \\
          & wood  & 99.8±0.3 & 99.9±0.1 & 98.9±0.5 & 96.4±0.3 & 95.7±0.3 & 69.1±1.1 \\
          & zipper & 99.5±0.2 & 99.9±0.1 & 98.7±0.4 & 98.3±0.4 & 94.2±0.5 & 54.7±0.6 \\
\cmidrule{2-8}          & \textbf{Mean} & \textbf{97.1±0.6} & \textbf{98.0±0.5} & \textbf{96.4±0.2} & \textbf{97.0±0.0} & \textbf{93.5±0.2} & \textbf{60.1±0.1} \\
    \midrule
    \midrule
    \multicolumn{1}{c}{\multirow{13}[4]{*}{\textbf{VisA}}} & candle & 94.0±0.4 & 94.4±0.6 & 87.5±0.7 & 99.1±0.1 & 96.8±0.1 & 38.2±0.6 \\
          & capsules & 97.5±0.2 & 98.4±0.1 & 94.5±0.6 & 98.1±0.4 & 96.8±0.5 & 48.6±4.6 \\
          & cashew & 92.7±1.5 & 96.7±0.6 & 89.6±1.8 & 98.7±0.2 & 98.0±0.5 & 59.1±0.2 \\
          & chewinggum & 97.6±0.1 & 99.0±0.0 & 95.5±0.3 & 99.2±0.1 & 85.7±1.2 & 68.1±0.8 \\
          & fryum & 95.8±0.9 & 98.2±0.4 & 94.0±1.2 & 96.2±0.6 & 90.8±0.6 & 41.7±0.7 \\
          & macaroni1 & 89.5±2.1 & 88.8±1.9 & 83.5±0.9 & 99.4±0.1 & 93.6±0.5 & 26.5±1.2 \\
          & macaroni2 & 66.2±5.0 & 63.2±4.8 & 70.0±1.1 & 98.5±0.1 & 87.6±0.2 & 22.3±1.8 \\
          & pcb1  & 95.4±0.7 & 93.3±0.8 & 92.7±0.8 & 98.9±0.1 & 94.2±0.1 & 60.4±2.0 \\
          & pcb2  & 88.4±1.1 & 85.8±0.8 & 82.9±0.6 & 95.7±0.1 & 86.5±0.2 & 38.5±1.1 \\
          & pcb3  & 89.0±1.4 & 91.6±1.1 & 83.1±2.0 & 95.2±0.5 & 87.4±0.3 & 42.5±3.9 \\
          & pcb4  & 91.0±3.5 & 91.9±2.7 & 86.1±4.0 & 95.3±0.3 & 83.1±1.3 & 30.3±1.0 \\
          & pipe\_fryum & 97.1±1.2 & 98.6±0.6 & 93.7±1.3 & 98.9±0.1 & 97.7±0.0 & 51.3±1.0 \\
\cmidrule{2-8}          & \textbf{Mean} & \textbf{91.2±1.1} & \textbf{91.7±0.8} & \textbf{87.8±0.9} & \textbf{97.8±0.0} & \textbf{91.5±0.2} & \textbf{44.0±0.6} \\
    \midrule
    \midrule
    \textbf{BraTS} & \textbf{brain} & \textbf{84.9±1.7} & \textbf{97.3±0.4} & \textbf{94.1±0.1} & \textbf{97.1±0.4} & \textbf{80.7±0.5} & \textbf{54.7±2.8} \\
    \midrule
    \midrule
    \multicolumn{1}{c}{\multirow{11}[4]{*}{\textbf{MVTec3D}}} & bagel & 96.0±1.9 & 99.0±0.5 & 95.8±1.4 & 99.0±0.1 & 99.7±0.0 & 67.0±3.2 \\
          & cable\_gland & 91.5±0.7 & 97.9±0.1 & 92.9±1.5 & 98.1±0.2 & 99.2±0.1 & 56.3±0.4 \\
          & carrot & 90.1±1.1 & 97.7±0.3 & 93.8±0.4 & 99.3±0.1 & 99.8±0.0 & 57.7±0.5 \\
          & cookie & 85.3±2.1 & 95.7±0.6 & 89.9±0.1 & 93.0±0.1 & 97.7±0.0 & 58.8±1.1 \\
          & dowel & 83.4±4.9 & 95.4±2.1 & 90.3±1.0 & 95.8±0.6 & 99.1±0.2 & 46.0±6.4 \\
          & foam  & 79.9±5.0 & 95.1±1.2 & 90.1±1.0 & 79.7±0.7 & 92.7±0.3 & 43.9±1.4 \\
          & peach & 96.8±1.5 & 99.2±0.3 & 96.3±1.2 & 99.4±0.0 & 99.8±0.0 & 69.0±0.6 \\
          & potato & 68.7±3.2 & 90.9±0.7 & 89.6±0.2 & 98.2±0.2 & 99.6±0.0 & 43.9±1.7 \\
          & rope  & 96.8±1.2 & 98.8±0.5 & 95.6±1.3 & 97.7±0.1 & 99.4±0.0 & 50.8±0.4 \\
          & tire  & 80.4±4.8 & 94.2±1.3 & 88.6±1.3 & 93.2±0.2 & 98.5±0.1 & 48.5±0.4 \\
\cmidrule{2-8}          & \textbf{Mean} & \textbf{86.9±1.8} & \textbf{96.4±0.5} & \textbf{92.3±0.7} & \textbf{95.4±0.0} & \textbf{98.6±0.0} & \textbf{54.2±0.4} \\
    \midrule
    \midrule
    \multicolumn{1}{c}{\multirow{6}[2]{*}{\textbf{MVTecLOCO}}} & breakfast\_box & 79.5±2.2 & 89.1±1.0 & 79.5±1.7 & 64.9±1.1 & 77.7±1.0 & 30.6±0.8 \\
          & juice\_bottle & 75.0±7.4 & 88.5±5.4 & 84.1±0.6 & 70.2±0.2 & 85.2±0.2 & 28.3±0.6 \\
          & pushpins & 65.0±8.2 & 69.1±5.8 & 74.5±3.0 & 53.6±1.1 & 53.5±2.0 & 3.4±0.3 \\
          & screw\_bag & 65.5±5.3 & 79.3±2.5 & 79.0±1.4 & 70.0±0.9 & 64.7±0.2 & 10.9±0.1 \\
          & splicing\_connectors & 72.7±9.1 & 81.6±8.0 & 78.5±1.7 & 76.1±1.4 & 65.2±4.0 & 19.9±1.3 \\
          & \textbf{Mean} & \textbf{71.5±3.1} & \textbf{81.5±2.2} & \textbf{79.1±0.8} & \textbf{66.9±0.6} & \textbf{69.2±0.5} & \textbf{18.6±0.4} \\
    \midrule
    \midrule
    \multicolumn{1}{c}{\multirow{4}[4]{*}{\textbf{BTAD}}} & 01    & 90.2±2.7 & 96.4±0.9 & 89.0±1.7 & 67.8±0.8 & 93.8±0.7 & 52.9±0.9 \\
          & 02    & 93.4±0.9 & 98.9±0.2 & 95.7±0.2 & 64.3±1.7 & 96.9±0.1 & 64.9±1.6 \\
          & 03    & 99.7±0.1 & 96.3±1.0 & 92.7±2.2 & 97.2±0.2 & 99.5±0.0 & 65.7±0.9 \\
\cmidrule{2-8}          & \textbf{Mean} & \textbf{94.4±1.2} & \textbf{97.2±0.5} & \textbf{92.5±0.9} & \textbf{76.4±0.9} & \textbf{96.8±0.2} & \textbf{61.2±0.4} \\
    \midrule
    \midrule
    \multicolumn{1}{c}{\multirow{7}[4]{*}{\textbf{MPDD}}} & bracket\_black & 68.9±11.5 & 76.0±10.0 & 78.9±2.2 & 95.4±1.6 & 97.6±1.2 & 27.2±3.0 \\
          & bracket\_brown & 60.7±8.8 & 73.3±5.4 & 83.2±1.1 & 90.4±1.7 & 96.0±0.5 & 15.3±2.0 \\
          & bracket\_white & 72.6±17.4 & 69.7±20.9 & 82.2±6.2 & 96.5±1.4 & 99.8±0.1 & 28.2±3.6 \\
          & connector & 86.3±2.7 & 74.6±3.5 & 77.2±3.9 & 92.2±1.1 & 97.7±0.3 & 36.2±6.0 \\
          & metal\_plate & 100.0±0.0 & 100.0±0.0 & 99.5±0.4 & 90.4±0.5 & 96.7±0.4 & 73.0±1.9 \\
          & tubes & 95.8±0.8 & 98.3±0.3 & 93.8±1.1 & 96.1±0.3 & 99.0±0.1 & 66.1±0.1 \\
\cmidrule{2-8}          & \textbf{Mean} & \textbf{80.7±3.0} & \textbf{82.0±3.9} & \textbf{85.8±1.3} & \textbf{93.5±0.7} & \textbf{97.8±0.3} & \textbf{41.0±1.1} \\
    \bottomrule
    \end{tabular}%
\end{table}%

\end{document}